\documentclass[lettersize,journal]{IEEEtran}
\usepackage{amsmath,amsfonts}
\usepackage{algorithmic}
\usepackage{algorithm}
\usepackage{array}
\usepackage[caption=false,font=normalsize,labelfont=sf,textfont=sf]{subfig}
\usepackage{textcomp}
\usepackage{stfloats}
\usepackage{url}
\usepackage{verbatim}
\usepackage{graphicx}
\usepackage{cite}
\usepackage{graphicx}
\usepackage{amsmath}
\usepackage{amssymb}
\usepackage{booktabs}
\usepackage{times}
\usepackage{graphicx}
\usepackage{amsmath}
\usepackage{amssymb}
\usepackage{color}
\usepackage{graphics}
\usepackage{multirow}
\usepackage{array}
\usepackage{colortbl}
\usepackage{cite}
\usepackage{diagbox}
\usepackage{rotating}
\usepackage{booktabs}
\usepackage{overpic}
\usepackage{contour}
\usepackage{bbding}
\usepackage[marginal]{footmisc}
\usepackage{enumitem}

\definecolor{mygray}{gray}{.92}

\newcommand{\figref}[1]{Fig. \ref{#1}}
\newcommand{\tabref}[1]{Tab. \ref{#1}}
\newcommand{\secref}[1]{$\S$\ref{#1}}

\newcommand{\tabincell}[2]{\begin{tabular}{@{}#1@{}}#2\end{tabular}}

\newcommand{\ourmodel}{HitNet}

\graphicspath{{./Imgs/}}
\DeclareGraphicsExtensions{.jpg,.pdf,.png}

\usepackage[pagebackref,breaklinks,colorlinks]{hyperref}

\usepackage[capitalize]{cleveref}
\crefname{section}{Sec.}{Secs.}
\Crefname{section}{Section}{Sections}
\Crefname{table}{Table}{Tables}
\crefname{table}{Tab.}{Tabs.}

\begin{document}

\title{High-resolution Iterative Feedback Network for Camouflaged Object Detection}

\author{Xiaobin Hu, Shuo Wang, Xuebin Qin, Hang Dai, Wenqi Ren, Ying Tai, Chengjie Wang, Ling Shao
\thanks{Xiaobin Hu, Ying Tai and Chengjie Wang  are with Tencent Youtu Lab, Shanghai, China.}%
\thanks{Shuo Wang is with CVL, ETH Zurich, Zurich, Switzerland.}%
\thanks{Hang Dai, Xuebin Qin and Ling Shao  are with Mohamed bin Zayed University of Artificial Intelligence
Abu Dhabi, United Arab Emirates.}%
\thanks{ Wenqi Ren is with the School of Cyber Science and Technology,
Sun Yat-sen University at Shenzhen, Shenzhen 518107, China}%
\thanks{Corresponding author: Hang Dai (hang.dai@mbzuai.ac.ae).}
}



\markboth{Journal of \LaTeX\ Class Files,~Vol.~14, No.~8, August~2021}%
{Shell \MakeLowercase{\textit{et al.}}: A Sample Article Using IEEEtran.cls for IEEE Journals}


\maketitle

\begin{abstract}
Spotting camouflaged objects that are visually assimilated into the background is tricky for both object detection algorithms and humans who are usually confused or cheated by the perfectly intrinsic similarities between the foreground objects and the background surroundings. To tackle this challenge, we aim to extract the high-resolution texture details to avoid the detail degradation that causes blurred vision in edges and boundaries. We introduce a novel \textbf{\ourmodel}
to refine the low-resolution representations by high-resolution features in an iterative feedback manner, essentially a global loop-based connection among the multi-scale resolutions. 
In addition, an iterative feedback loss is proposed to impose more constraints on each feedback connection.
Extensive experiments on four challenging datasets demonstrate that our \ourmodel~breaks the performance bottleneck and achieves significant improvements compared with 29 state-of-the-art methods. 
To address the data scarcity in camouflaged scenarios,
%
%
we provide an application example by employing 
the cross-domain learning 
to extract the features that can reflect the camouflaged object properties and embed the features into salient objects, thereby generating more camouflaged training samples from the diverse salient object datasets.
The code will be available at: \textit{https://github.com/HUuxiaobin/HitNet}.
%
\end{abstract}


\section{Introduction}
Camouflaged object detection (COD) is a bio-inspired research area to detect hidden objects or animals that blend with their surroundings~\cite{fangtpami2021}. From biological and psychological studies~\cite{cuthill2019camouflage,stevens2009animal}, the camouflage skill helps some animals prevent being the prey of their predators, and it also can cheat the human perception system that is sensitive to the coloration and the illumination around the edges. The camouflaged studies not only provide an effective way to deeply understand human perception system, but also benefit a wide range of downstream applications, such as medical image segmentation \cite{dong2021PolypPVT,fan2020pranet,fan2020inf}, artistic creation \cite{chu2010camouflage}, species discovery \cite{perez2012early}, and crack inspection \cite{fang2020novel}. 

In the last two decades, a growing interest is witnessed in developing algorithms capable of seeing targets through camouflage. Early methods aim to utilize the handcrafted low-level features (\textit{e.g.,} texture and contrast \cite{huerta2007improving}, 3D convexity \cite{pan2011study} and motion boundary \cite{hou2011detection}). These features still suffer from the limited capability of discriminating the foreground and the background in complex scenes.
Recently, some CNNs-based frameworks have been proposed to analyze the visual similarities around boundaries between the camouflaged objects and their surroundings. The auxiliary information is extracted from the shared context as the boundary guidance for COD, such as features for identification \cite{fan2020camouflaged}, classification \cite{le2019anabranch}, boundary detection \cite{zhai2021mutual} and uncertainties \cite{yang2021uncertainty}.

\begin{figure}[t!]
\centering
\includegraphics[width=.99\columnwidth]{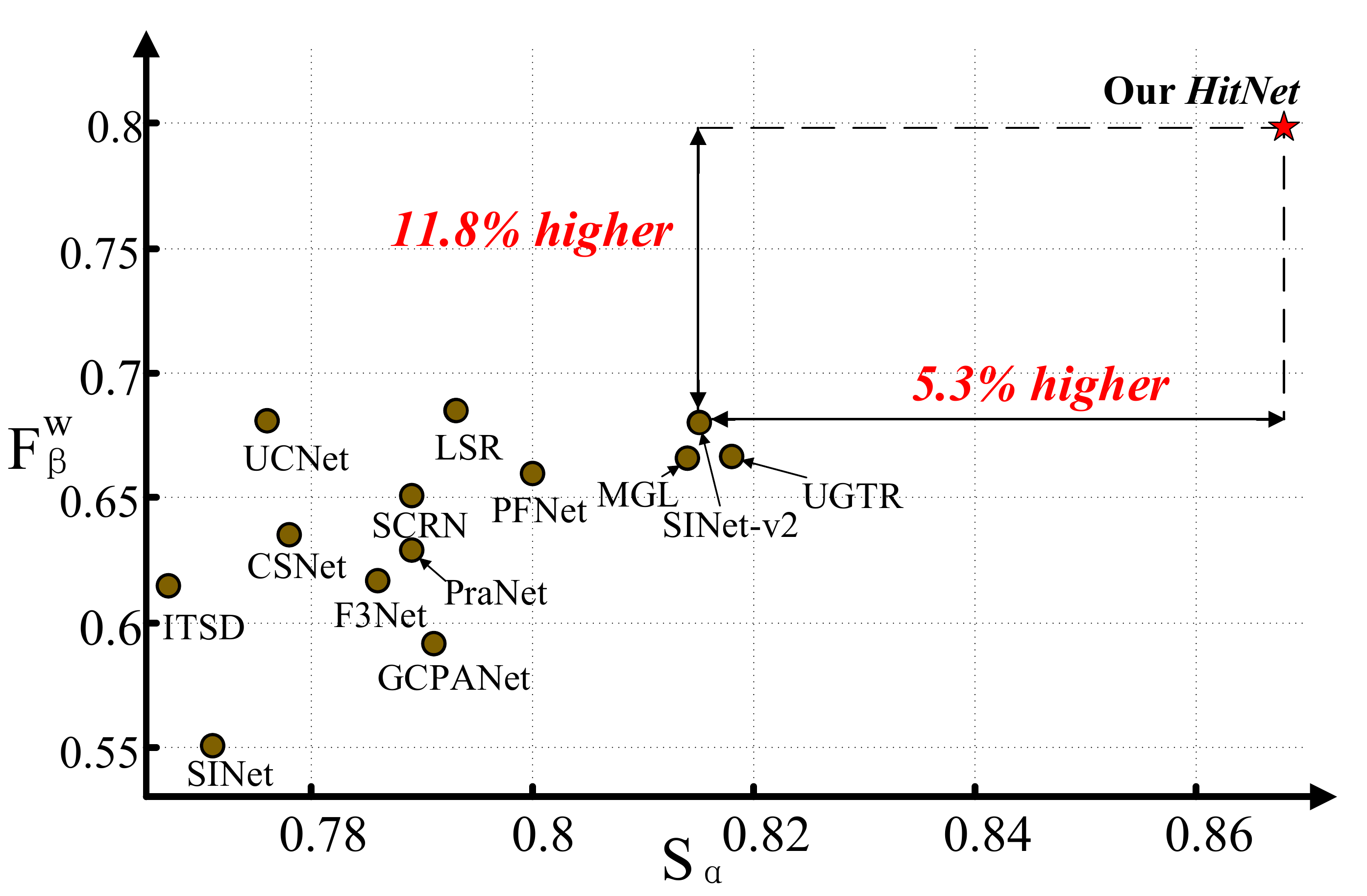}
\caption{\small 
Weighted F-measure ($F_\beta^w$) \textit{vs.} Structure-measure ($S_{\alpha}$) of top 13 models from 29 SOTA methods and our \textit{\ourmodel} on COD10k-Test dataset. $F_\beta^w$ is a comprehensive metric to evaluate the weighted precision and recall of the prediction map, and $S_{\alpha}$ aims to analyze the structural information of the prediction map. Our framework achieves a remarkable performance milestone. 
}\label{fig:fig1_results}
\end{figure}

\begin{figure*}[t!]
	\centering
    \small
	\begin{overpic}[width=.98\textwidth]{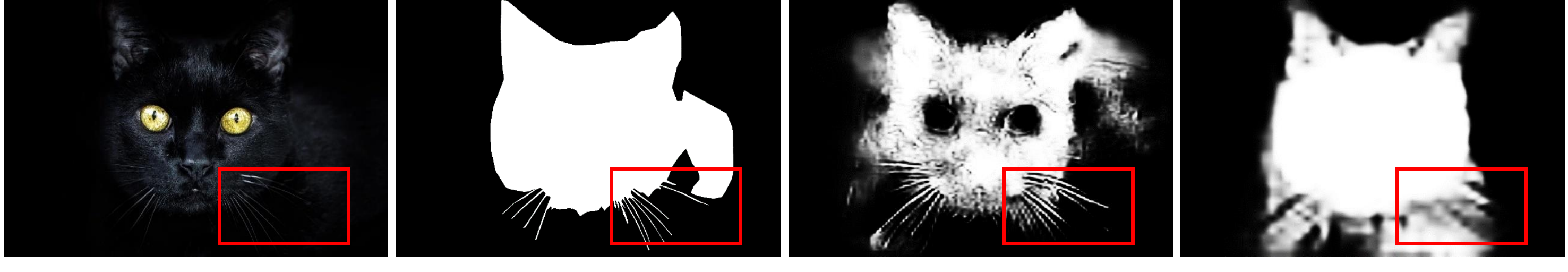} %
    \put(9,-1) {(a) Image}
    \put(31,-1){(b) Ground-truth}
    \put(58,-1){(c) HR result}
    \put(82,-1){(d) LR result}
    \end{overpic}
	\caption{Results from high-resolution (HR) and low-resolution (LR) inputs with SINet~\cite{fan2020camouflaged} trained on LR images. LR result has blurred edges (\textit{e.g.,} cat's beard), which indicates the details loss (\textit{e.g.,} boundaries) during image degradation process from HR to LR.
    }
    \label{fig:Myopia-Symptom}
\end{figure*}
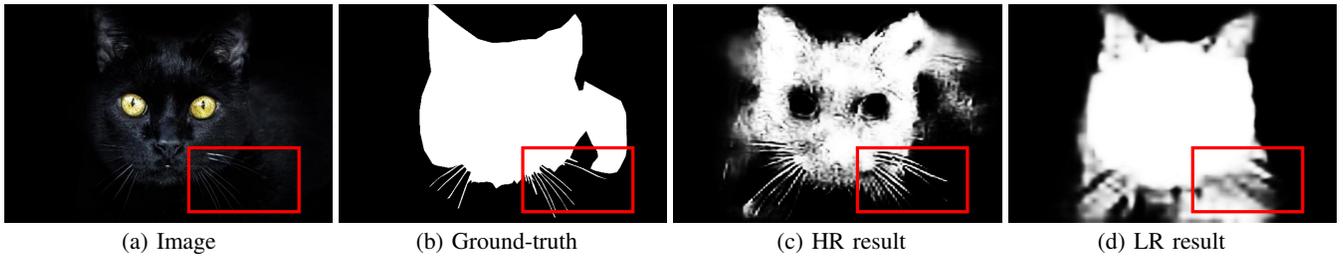

Although the approaches mentioned above have improved the performance, most methods discard the high-resolution details, including edges or textures, by down-sampling the high-resolution images. 
\figref{fig:Myopia-Symptom} shows an interesting phenomenon by evaluating the low-resolution (LR) and the high-resolution (HR) images on the same model well-trained on LR images, respectively. The segmentation result from HR has more details like cat beards than that from LR. 
This implies that the high-resolution priors are crucial to the boundary and edge detection \cite{zhang2021looking,wang2021tpami}. 
The degradation of inputs from HR to LR leads to blurry vision without capturing fine structures.
To balance the trade-off between computational resources and performance, the down-sampling operation on high-resolution input is acceptable to achieve satisfactory performance to some extent. 
But the lose of edge details is not desirable in segmentation tasks, especially for camouflaged object detection. 
%
We find that two main aspects account for the degradation: 1) the lack of high-resolution information from input images; 2) the absence of an effective mechanism to enhance the low-resolution features. 
Thus, it is promising to explore how to maintain the high-resolution information at input level and enhance the low-resolution features without sacrificing the real-time property. 

To achieve this goal, we propose a High-resolution Iterative Feedback Network (\textbf{\ourmodel}) to sufficiently and comprehensively exploit the multi-scale HR information and refine the LR with HR knowledge via an adaptively iterative feedback approach. Specifically, \ourmodel~includes three main components: Transformer-based Feature Extraction (TFE),Multi-resolution iterative refinement (RIR), and Iteration feature feedback (IFF). 
To reduce the computational cost for the HR feature maps, we adopt pyramid vision transformer~\cite{Wang_2021_ICCV} as an image feature encoder through a progressive shrinking pyramid and spatial-reduction attention. Then, we utilize the RIR module to recursively refine the LR feature extracted from TFE via a global and cross-scale feedback strategy. 
To ensure the better aggregation of feedback feature, we use iteration feature feedback (IFF) to impose constraints on feedback feature flow.

%
In addition, we implement an application that converts the salient objects to camouflaged objects via our cross-domain learning strategy. 
Results from our application can be used as additional training data for the COD task without increasing the parameters and computations of deep learning models in the inference stage. 

Our main contributions are summarized as:
\begin{itemize}
    \item We propose a novel recursive operation to refine the low-resolution feature via a cross-scale feedback mechanism. 
    The recursive operation is simple and can be easily extended to existing COD models. 
    \item Based on the recursive operation, we design a novel framework, termed as High-resolution Iterative Feedback Network (\textbf{\ourmodel}) for COD task. The corresponding iterative feedback loss with an iteration weight scheme is also proposed for \textbf{\ourmodel} to penalize the output of each iteration.
    \item Our \ourmodel~sets a new record, as shown in \figref{fig:fig1_results}, breaking the performance bottleneck, compared with existing cutting-edge models on 
    four benchmarks using four standard metrics. On COD10K, \ourmodel~achieves $F_\beta^w$ of 0.798, which is \textbf{16.5}\% higher than the second-best LSR \cite{lv2021simultaneously}. On CHAMELEON, our \ourmodel~achieves a mean MAE error of 0.018, which is \textbf{40.0}\% better than the second-best SINet-v2 \cite{fangtpami2021}.
\end{itemize}

\section{Related Work}\label{sec:Related Work}
\noindent\textbf{Camouflaged Object Detection.} COD aims to spot the camouflaged object from its high-similarity surroundings \cite{stevens2008animal, stevens2011animal, owens2014camouflaging, fan2020camouflaged}.  
It has wide applications \cite{fan2020pranet,chu2010camouflage,perez2012early} and many COD methods \cite{youwei2022zoom,cheng2022implicit} have been proposed. These methods can be categorized into two main classes: handcrafted-based and deep-learning-based. 
More specifically, most of the early works were developed based on the handcrafted features (\textit{e.g.,} colour and intensity features \cite{huerta2007improving}, 3D convexity \cite{pan2011study}, and motion boundary \cite{hou2011detection}). But they are relatively less robust and prone to fail in complex scenarios. 
More studies resort to the powerful representation capacity of deep learning models to detect camouflaged objects in a data-driven way and have achieved impressive improvements against those handcrafted-based methods. On the one hand, deep models usually have many parameters, which ensures stronger representative capabilities for segmenting the camouflaged objects from their backgrounds. On the other hand, most of these deep models benefit by exploring the auxiliary knowledge, \textit{e.g.,} fixations \cite{lv2021simultaneously}, boundaries \cite{zhai2021mutual}, location \cite{fan2020camouflaged}, image-level labels \cite{le2019anabranch}, and uncertainty analysis \cite{yang2021uncertainty}. Nevertheless, most of models pay much attention to regional accuracy. At the same time, few of them explore the effectiveness of high-frequency information (in high-resolution), which plays a vital role in perceiving the clear boundaries or edges of camouflaged targets. Thus, it impedes the further improvements of COD models. 
To address this issue, we design a novel High-resolution Iterative Feedback Network, which sets a new record on all benchmarks.

\begin{figure*}[t!]
\centering
\includegraphics[width=.95\textwidth]{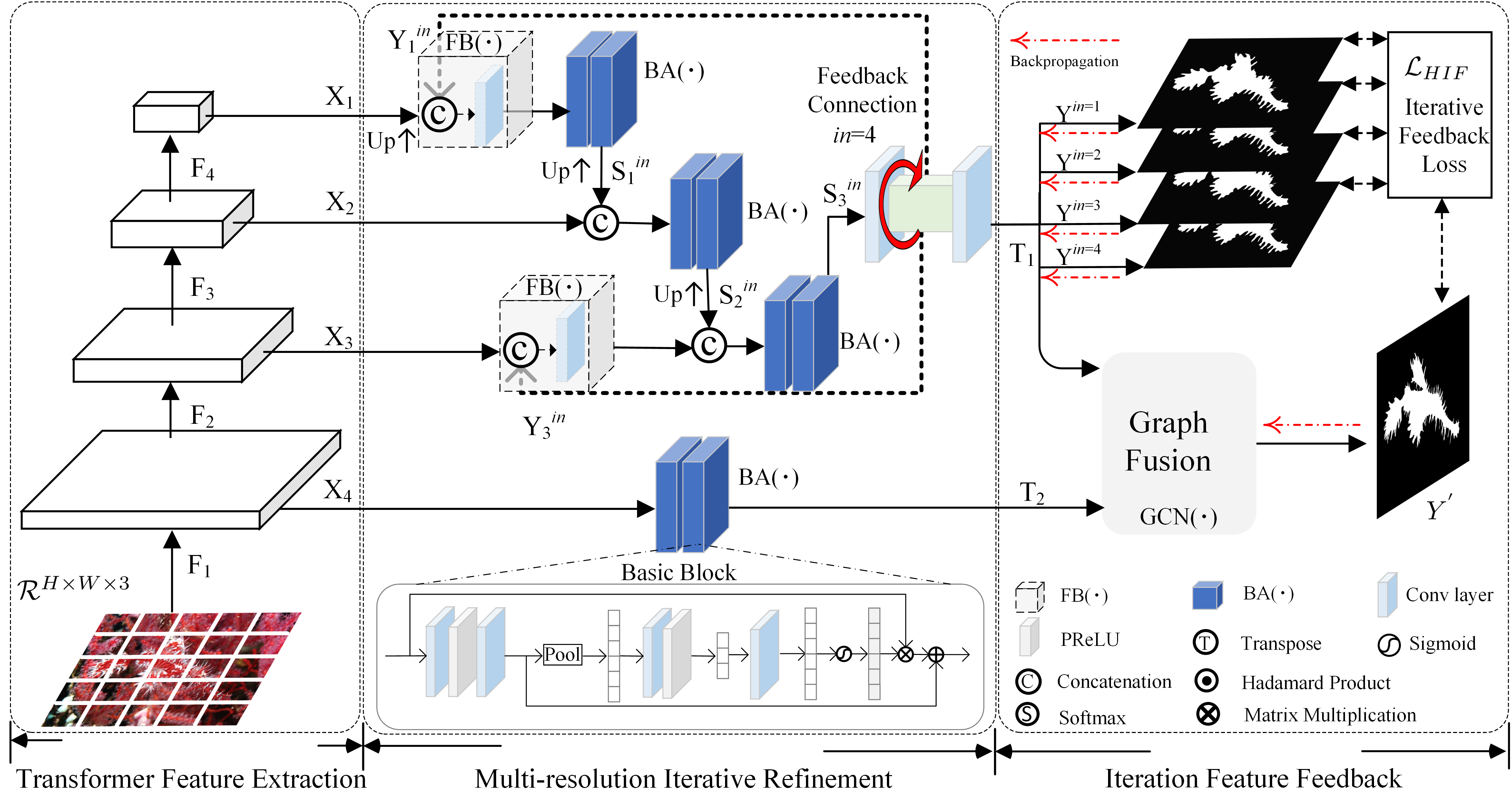}
\caption{An overview of Transformer-based High-resolution Iterative Feedback Network (\textit{\ourmodel}). Our \ourmodel~consists of Transformed-based backbone for multi-scale feature extraction,multi-resolution iterative refinement to self-correct low-resolution features with high-resolution information via a cross-resolution iterative feedback mechanism, 
and iteration feature feedback to impose constraint on each iteration. 
}\label{fig:fig3_net}
\end{figure*}

\noindent\textbf{Iterative Feedback Mechanism of Super-Resolution} allows the network to correct previous states (\textit{i.e.,} lower-resolution) with a higher-level output (\textit{i.e.,} higher-resolution) \cite{zamir2017feedback, hu2021feedback}. 
In image super-resolution, some studies proved certain improvements after using different feedback mechanisms, such as up- and down-projection units \cite{haris2018deep} and dual-state recurrent module \cite{han2018image}. However, most of these mechanisms are implemented by using recurrent structures \cite{li2019feedback} while the information flows from the LR to HR images are still feed-forward. 
Recently, Li \textit{et al.} \cite{li2019feedback} proposed an image super-resolution feedback network to refine LR representation with HR information by outlining the edges and contours while suppressing smooth areas. Inspired by this work, we build our transformer-based high-resolution iterative feedback for COD. Different from Li \textit{et al.} \cite{li2019feedback}, our feedback connection is designed as a global connection other than a local connection \cite{feng2019attentive} and embedded into the multi-scale framework via a feedback fusion block, which merges the information from multi-scale outputs. 
To avoid the corruption of each iteration, we impose more constraints on each feedback connection by supervising each iteration with the corresponding loss. 

\noindent\textbf{Vision Transformer.} The Transformer was firstly proposed by Vaswani \textit{et al.} \cite{vaswani2017attention} as a powerful tool in the domain of machine translation. Considering the superiority of transformer in modeling long-term dependencies, more recent studies have tried to exploit its potentials in different vision tasks, such as image classification \cite{dosovitskiy2020image,srinivas2021bottleneck}, semantic segmentation \cite{zheng2021rethinking}, object detection \cite{dai2021up}, and other low-level tasks \cite{yang2020learning}. Those works verify the image data can be learned in a sequence-to-sequence approach. Unlike the convolution layer, the layer of multi-head self-attention in the Transformer has dynamic weights and a global receptive field, making the Transformer more effective and powerful in catching non-local knowledge. 
However, the Transformer suffers high computational and memory costs, especially for HR inputs. Thus, we adopt a Pyramid Vision Transformer (PVT) \cite{Wang_2021_ICCV} that uses a progressive shrinking pyramid structure to reduce the sequence length and a spatial-reduction attention layer to decrease the computation further when learning HR features.

\section{Proposed Method}
\noindent\textbf{Motivation.}
Our motivation stems from the observation of degradation phenomenon, shown in \figref{fig:Myopia-Symptom}, HR inputs generate more accurate predictions than LR inputs, especially for object boundaries. 
Thus, we aim to explore 
the feature interaction between high- and low-resolution for COD.

\noindent\textbf{Overview.} To achieve this goal, we design \textit{\ourmodel}, consisting of three main modules: Transformer-based Feature Extraction (Sec.~\ref{sec:TFE}), Multi-resolution iterative refinement (Sec.~\ref{section:3parts}), and
Iteration Feature Feedback
(Sec.~\ref{sec:IFF}).

\subsection{Transformer-based Feature Extraction}\label{sec:TFE}


Recently, the transformer \cite{vaswani2017attention}, which was originally developed for NLP tasks, are proved to be very competitive in many vision tasks against the existing CNN-based backbones. However, many of the vision transformers are GPU memory exhaustive and our HR features will further exaggerate the problem. To alleviate this issue, we choose the Pyramid Vision Transformer (PVT) \cite{Wang_2021_ICCV} as our feature extraction module, which can extract multi-scale features, and handle relatively higher resolution feature maps with less memory costs by its progressive shrinking strategy and spatial reduction attention mechanism. 

\noindent\textbf{Progressive Shrinking Strategy:}  Denoting the patch size of the $i$-th stage as $P_i$.
Input feature $F_{i-1} \in \mathcal{R} ^{H _{i-1}\times W_{i-1} \times C_{i-1}}$ will be evenly split into $\frac{H_{i-1}\times W_{i-1}}{P_{i}\times P_{i}}$ patches. After the linear projection, the height and width are $\frac{1}{P_i}$ of the input features. 

\noindent\textbf{Spatial Reduction Attention:} 
PVT employs spatial reduction attention (SRA) to decrease the computational cost. 
More specifically, after receiving a query $Q$, a key $K$, and a value $V$, SRA reduces the spatial dimension of $K$ and $Q$ before the attentive operation. 
Given a input sequence $x \in \mathcal{R}^{H_{i}W_{i}\times C_{i}}$ at $i$-th stage, 
a reduction ratio of spatial dimension $R_{i}$ is used to reshape the input $x$ to the size $\frac{H_{i}W_{i}}{R^{2}_{i}}\times(R^{2}_{i}C_{i})$.
Then, a linear projection is adopted to further reduce the dimension from input sequence (\textit{i.e.,} $R^{2}_{i}C_{i}$) to $C_{i}$. Consequently, the computational cost of SRA only occupies $\frac{1}{R^{2}_{i}}$ of the standard Multi-head attention layer~\cite{vaswani2017attention}.

\noindent\textbf{Multi-scale Feature Extraction:} PVT consists of four stages, and each stage includes a patch embedding and an encoder structure. The input features to each stage ($\boldsymbol{F_{i}}$) are first divided into patches with size of $P_i$. After that, these features are fed into Transformer encoder structure to get the output features $X_i$ for the $i$-th. Then, we get the multi-scale features ($\boldsymbol{X_{1}}$, $\boldsymbol{X_{2}}$, $\boldsymbol{X_{3}}$ ,$\boldsymbol{X_{4}}$) with ($\frac{1}{32}$, $\frac{1}{16}$, $\frac{1}{8}$, $\frac{1}{4}$) resolution of input images for further processing.
\subsection{multi-resolution feedback refinement}\label{section:3parts}
The multi-scale resolution features $\boldsymbol{X}$ extracted from the Transformer backbone are fed to a basic block ${\rm BA}(\cdot)$~\cite{zhang2018image} as shown in Fig. \ref{fig:fig3_net}:
\begin{equation}
\text{BA}(\boldsymbol{X_{i}})=C_{2}\boldsymbol{({X_{i}})}+C_b(C_{2}(\boldsymbol{X_{i}}))\cdot\boldsymbol{X_{i}},
\end{equation}
where $\boldsymbol{X_{i}}$ is the input feature of $i$-th scale produced by Transformer module, $C_2(\cdot)$ indicates two stacked convolutional layers with $3\times3$ filters. $C_b(\cdot)$ denotes the channel attention function~\cite{zhang2018image}.


\textbf{Iterative Feedback Mechanism} is critical in this module to achieve high accuracy around the object boundary (see \figref{fig:fig6_itnum}). 
The setting iterative number \textit{in}=1 assumes the first iteration and no feedback feature transported from previous state. Thus, the $\boldsymbol{Y^{in}_{1}}$ and $\boldsymbol{Y^{in}_{1}}$ are the initial value (0) when \textit{in}=1.
For iterative number ($in>1$), the feedback features are produced by previous iteration and then passed into feedback block ${\rm FB}(\cdot)$ as: 
\begin{equation}
{\rm FB}(\boldsymbol{X_{i}+Y^{in}_{i}})={\rm Sq}({\rm Concat}(\boldsymbol{X_{i}}\uparrow, \boldsymbol{Y^{in}_{i}})),
\end{equation}
where $\boldsymbol{Y^{in}_{i}}$ is the feedback features of $in$-th iteration at $i$-th scale ($i\neq2$), Symbol $\uparrow$ is the up-sampling operation from the size of $\boldsymbol{X_{i}}$ to $\boldsymbol{Y^{in}_{i}}$ to avoid degradation of the HR information. ${\rm Concat(\cdot)}$ indicates the channel-based concatenation operation between $\boldsymbol{X_{i}}\uparrow$ and $\boldsymbol{Y^{in}_{i}}$, and ${\rm Sq(\cdot)}$ is size and channel compression using the convolution layer with large kernel and stride\footnote{If i=1, kernel = 8 with stride = 4 while i=3, kernel = 1 with stride = 1.} to obtain identical size for $i$-th scale. 

As shown in Fig. \ref{fig:fig3_net}, with the prerequisite that the iterative number ($in>1$), the first scale structure receives $\boldsymbol{X_{1}}$ and $\boldsymbol{Y^{in}_{1}}$ and the output the feature can be defined as:
\begin{equation}
\boldsymbol{S_{1}^{in}}={\rm BA}({\rm FB}(\boldsymbol{X_{1}+Y^{in}_{1}})).
\end{equation}
Then, $\boldsymbol{S_{1}^{in}}$ is further fed into the next scale to generate next output feature as follows:
\begin{equation}
\boldsymbol{S_{2}^{in}}={\rm BA}({\rm Concat}(\boldsymbol{S_{1}^{in}}\uparrow, \boldsymbol{X_{2}})),
\end{equation}
Finally, the features of the previous scale are transported to the next scale as:
\begin{equation}
\boldsymbol{S_{3}^{in}}={\rm BA}({\rm Concat}(\boldsymbol{S_{2}^{in}}\uparrow,{\rm FB}(\boldsymbol{X_{3}+Y^{in}_{3}}))).
\end{equation}

After ending at $in$-th iteration, $(in+1)$-th iteration starts from the first scale to the last scale in the same way. 
The design intuitions on different scales are mainly motivated to get a better cross-scale data flow. The feedback features are explicitly imported into the top and third top scales for the data flow. 
As the data flow works, the second-top scale can get the implicit feedback features from the top scale. From our experiments, this setting can decrease the computational cost
but maintaining good performance. 
Our \ourmodel~breaks the performance bottleneck due to the following three indispensable mechanisms:
\begin{itemize}
    \item In each iteration, it outputs an intermediate HR segmentation map that is supervised with a segmentation loss, enabling the feedback features to learn HR cues.
    \item The HR feedback features merge with inputs features in a feedback block, alleviating the degradation of HR information. 
     \item It uses a feedback fusion mechanism to exploit the HR data flow in a multi-scale structure.
\end{itemize}

\subsection{Iteration Feature Feedback}\label{sec:IFF}
To tailor satisfactory feedback feature flow and avoid the feature corruption caused by recurrent path, we present iteration feature feedback strategy to tie the each feedback feature with the segmentation ground-truth. Intuitively, the data flow of feedback features can be controlled by the loss function.
Our basic loss function is defined as $\mathcal{L}=\mathcal{L}^{w}_{IoU}+\mathcal{L}^{w}_{BCE}$, where $\mathcal{L}^{w}_{IoU}$ is the weighted intersection-over-union (IoU) loss and $\mathcal{L}^{w}_{BCE}$ denotes the weighted binary cross entropy (BCE) loss. Unlike other recurrent structures \cite{wei2020f3net}, we compute the HR prediction loss in each iteration and use an iteration-weight scheme to penalize the output of each iteration when predicting a HR segmentation map: 
\begin{equation}
\mathcal{L}_{HIF}= \sum ^{N}_{in}(w \cdot in) \mathcal{L}(Y^{in})+ \mathcal{L}(Y^{'}),
\end{equation}
where $in$ is the current iteration number, $N$ is the total iteration number, $w$ is the weight parameter, $Y^{in}$ is the output of $in$-th iteration, $Y^{'}$ is the output of graph-based resolution fusion. 
In this way, our iteration-weight scheme focuses on the features of deeper iterations by assigning higher weights.

In this session, to efficiently integrate the features from the previous module, we introduce the non-local graph fusion (shown in Fig. \ref{fig:fig3_net}) via a graph fusion module~\cite{dong2021PolypPVT} and \cite{te2020edge}. 
\begin{equation}
Y^{'}={\rm GCN}(T_{1},T_{2}),
\end{equation}
where $Y^{'}$ is final prediction map, $T_{1}$ is the $Y^{in=4}$, $\rm GCN$ is the graph fusion module. For more details regarding the graph fusion, we
refer the readers to \cite{dong2021PolypPVT} and \cite{te2020edge}. 
.


\begin{table*}[t!]
  \footnotesize
  \caption{\small 
  Quantitative results of our method and other 29 state-of-the-art methods on four benchmark datasets. The best results are highlighted in \textbf{bold}, and the second-best is marked in \underline{underline}. Our \textit{\ourmodel} outperforms the second-best model by a large margin. 
  }\label{tab:ModelScore}
  \renewcommand{\arraystretch}{0.9} 
  \setlength\tabcolsep{3.7pt} 
  \begin{tabular}{r||cccc|cccc|cccc|cccc}
  \toprule
    & \multicolumn{4}{c|}{\tabincell{c}{CHAMELEON~\cite{2018Animal}}} & \multicolumn{4}{c|}{\tabincell{c}{CAMO-Test~\cite{le2019anabranch}}} & \multicolumn{4}{c|}{ \tabincell{c}{COD10K-Test \cite{fan2020camouflaged}}} & \multicolumn{4}{c}{ \tabincell{c}{NC4K \cite{lv2021simultaneously}}} \\
   \cline{2-17}
    Baseline Models~~~ &$S_\alpha\uparrow$      &$E_\phi\uparrow$     &$F_\beta^w\uparrow$      &$M\downarrow$
               &$S_\alpha\uparrow$      &$E_\phi\uparrow$     &$F_\beta^w\uparrow$      &$M\downarrow$
               &$S_\alpha\uparrow$      &$E_\phi\uparrow$     &$F_\beta^w\uparrow$      &$M\downarrow$
               &$S_\alpha\uparrow$      &$E_\phi\uparrow$     &$F_\beta^w\uparrow$      &$M\downarrow$
               \\
     \hline
     2017~MaskRCNN~\cite{he2017mask}
         &0.643&0.778&0.518&0.099&0.574&0.715&0.430&0.151&0.613&0.748&0.402&0.080 & $\ddagger$ & $\ddagger$ & $\ddagger$& $\ddagger$\\
     2017~FPN~\cite{lin2017feature}
         &0.794&0.783&0.590&0.075&0.684&0.677&0.483&0.131&0.697&0.691&0.411&0.075 & $\ddagger$ & $\ddagger$ & $\ddagger$& $\ddagger$\\
    2017~NLDF \cite{luo2017non}
    &0.798 &0.809 & 0.714  &0.063
    & 0.665& 0.664 & 0.564 &0.123 
    & 0.701& 0.709 & 0.539& 0.059 
    &0.738& 0.748 & 0.657 &0.083 \\
     2017~PSPNet~\cite{zhao2017pyramid}
         &0.773&0.758&0.555&0.085&0.663&0.659&0.455&0.139&0.678&0.680&0.377&0.080 & $\ddagger$ & $\ddagger$ & $\ddagger$& $\ddagger$\\
     2018~UNet++~\cite{zou2018DLMIA}
         &0.695&0.762&0.501&0.094&0.599&0.653&0.392&0.149&0.623&0.672&0.350&0.086 & $\ddagger$ & $\ddagger$ & $\ddagger$& $\ddagger$\\
     2018~PiCANet~\cite{liu2018picanet}
         &0.769&0.749&0.536&0.085&0.609&0.584&0.356&0.156&0.649&0.643&0.322&0.090 & 0.758 &0.773&  0.639 &0.088 \\
     2019~MSRCNN~\cite{huang2019mask}
         &0.637&0.686&0.443&0.091&0.617&0.669&0.454&0.133&0.641&0.706&0.419&0.073 & $\ddagger$ & $\ddagger$ & $\ddagger$& $\ddagger$\\
     2019~PoolNet~\cite{liu2019simple}
         &0.776&0.779&0.555&0.081&0.702&0.698&0.494&0.129&0.705&0.713&0.416&0.074 &0.785& 0.814 &0.699&  0.073 \\
         
     2019~BASNet~\cite{qin2019basnet}
         &0.687&0.721&0.474&0.118&0.618&0.661&0.413&0.159&0.634&0.678&0.365&0.105
         & 0.698 & 0.761 & 0.613 & 0.094 \\
         
      2019~SCRN \cite{wu2019stacked} 
     & 0.876& 0.889& 0.787 & 0.042 
     &0.779 &0.796& 0.705 & 0.090 
     &0.789& 0.817& 0.651 &0.047
     &0.832 &0.855 &0.759&  0.059 \\   
     2019~PFANet~\cite{zhao2019pyramid}
         &0.679&0.648&0.378&0.144&0.659&0.622&0.391&0.172&0.636&0.618&0.286&0.128 & $\ddagger$ & $\ddagger$ & $\ddagger$& $\ddagger$\\
     2019~CPD~\cite{wu2019cascaded}
         &0.853&0.866&0.706&0.052&0.726&0.729&0.550&0.115&0.747&0.770&0.508&0.059 & 0.790 &0.810 &0.708 & 0.071 \\
     2019~HTC~\cite{chen2019hybrid}
         &0.517&0.489&0.204&0.129&0.476&0.442&0.174&0.172&0.548&0.520&0.221&0.088 & $\ddagger$ & $\ddagger$ & $\ddagger$& $\ddagger$\\
     2019~EGNet~\cite{zhao2019EGNet}
         &0.848&0.870&0.702&0.050&0.732&0.768&0.583&0.104&0.737&0.779&0.509&0.056 &0.796& 0.830 &0.718 & 0.067
         \\
    2019~ANet-SRM~\cite{le2019anabranch}
         & $\ddagger$ & $\ddagger$ & $\ddagger$& $\ddagger$  &0.682&0.685&0.484&0.126& $\ddagger$ & $\ddagger$ & $\ddagger$ & $\ddagger$ & $\ddagger$ & $\ddagger$ & $\ddagger$& $\ddagger$\\     
   2020~CSNet \cite{gao2020highly} &0.856 &0.869& 0.766&  0.047
   &0.771 &0.795& 0.705&  0.092 
   &0.778& 0.810 &0.635 & 0.047 
   & 0.819& 0.845 &0.748 & 0.061 \\

    2020~MirrorNet~\cite{yan2020mirrornet}
                    & $\ddagger$ & $\ddagger$ & $\ddagger$& $\ddagger$ &0.741 &0.804 &0.652& 0.100& $\ddagger$ & $\ddagger$ & $\ddagger$ & $\ddagger$  & $\ddagger$ & $\ddagger$ & $\ddagger$& $\ddagger$  \\ 
    2020 PraNet~\cite{fan2020pranet}
                    &0.860 &0.898 &0.763& 0.044& 0.769& 0.833 &0.663& 0.094& 0.789& 0.839& 0.629& 0.045 & 0.822 & 0.876 & 0.724 & 0.059\\
                    
    2020 SINet~\cite{fan2020camouflaged}
                    &0.869 &0.891& 0.740 &0.044&0.751 &0.771& 0.606 &0.100& 0.771& 0.806& 0.551 &0.051     &  0.810 & 0.873 & 0.772 & 0.057 \\  
    2020 F3Net~\cite{wei2020f3net}
      &0.854  &0.899  &0.749 & 0.045 & 0.779 & 0.840  &0.666 & 0.091 & 0.786 & 0.832 & 0.617 & 0.046  & 0.782 &0.825 &0.706 &0.069 \\
    
    2020~UCNet \cite{zhang2020uc} & 0.880& 0.930 & {0.836} & 0.036 & 0.739 &0.787& 0.700 & 0.094  & 0.776 &0.857 &0.681 & 0.042 & 0.813& 0.872 &0.777 & 0.055 \\
      
    2020~ITSD \cite{zhou2020interactive}  &0.814 &0.844 &0.705 & 0.057 
    &0.750& 0.779 &0.663 & 0.102 
    &0.767& 0.808 &0.615& 0.051 
    &0.811& 0.845 &0.729&  0.064  \\ 

    2020~SSAL~\cite{zhang2020weakly}
    &0.757 &0.849 &0.702  &0.071 
    &0.644 &0.721 &0.579  &0.126 
    &0.668 &0.768 &0.527  &0.066 
    &0.699 &0.778 &0.647  &0.092 \\
    2020 GCPANet~\cite{chen2020global} 
    &0.876& 0.891 &0.748& 0.041& 0.778 &0.842& 0.646& 0.092& 0.791& 0.799& 0.592& 0.045 & $\ddagger$ & $\ddagger$ & $\ddagger$& $\ddagger$\\
    2021 MGL~\cite{zhai2021mutual}
    &\underline{0.893} &0.923& 0.813& \underline{0.030}& 0.775& 0.847& 0.673& 0.088& 0.814& 0.865& 0.666 &\underline{0.035} & $\ddagger$ & $\ddagger$ & $\ddagger$& $\ddagger$\\
    2021 PFNet~\cite{mei2021camouflaged} &0.882 &\underline{0.942} &0.810 &0.033& 0.782& 0.852& 0.695& 0.085& 0.800& 0.868& 0.660& 0.040 & 0.829 & 0.887 & 0.745 & 0.053 \\
     2021 UGTR~\cite{yang2021uncertainty}  &0.888 &0.918& 0.796& 0.031& 0.785& 0.859& 0.686 &0.086& \underline{0.818}& 0.850 &0.667& \underline{0.035} & $\ddagger$ & $\ddagger$ & $\ddagger$& $\ddagger$ \\

     2021~LSR~\cite{lv2021simultaneously}  & \underline{0.893} &  0.938 &  \underline{0.839} &  0.033 & 0.793 &  0.826 & 0.725 & 0.085 & 0.793 &  0.868 &  \underline{0.685} & 0.041 
     & 0.839 & 0.883 & \underline{0.779} & 0.053
     
     
     \\ 

    2021 SINet-V2~\cite{fangtpami2021} &0.888 & \underline{0.942} &0.816&\underline{0.030} &\underline{0.820}& \underline{0.882}& \underline{0.743}& \underline{0.070}& 0.815& \underline{0.887}& 0.680& 0.037 &
     \underline{0.847} & \underline{0.903} & 0.769 & \underline{0.048} \\
     2022 \textbf{\ourmodel~(Ours)} & \textbf{0.922} & \textbf{0.970} & \textbf{0.903} & \textbf{0.018} & \textbf{0.844} & \textbf{0.902} & \textbf{0.801} & \textbf{0.057}& \textbf{0.868} & \textbf{0.932} & \textbf{0.798} & \textbf{0.024} & \textbf{0.870} & \textbf{0.921} & \textbf{0.825} & \textbf{0.039} \\
    

  \toprule
  \end{tabular}
\end{table*}    


\section{Experiments}
\vspace{-5pt}
\subsection{Experimental Settings}
\vspace{-5pt}
\noindent\textbf{Datasets.} 
Our experiments are based on four widely-used COD datasets: 
(1) CHAMELEON~\cite{2018Animal} collects 76
high-resolution images from the Internet with the label of camouflaged animals. Each image is manually annotated with object-level GT masks.
(2) CAMO~\cite{le2019anabranch} includes 2500 images with eight categories, and 2000 images of them for training and 500 images for testing.
(3) COD10K \cite{fan2020camouflaged} is the largest collection containing 10,000 images that divided into 10 super-classes and 78 sub-classes from multiple photography websites.
(4) NC4K~\cite{lv2021simultaneously} consists of 4,121 images and is commonly used to evaluate the generalization ability of models.
Following previous studies\cite{zhai2021mutual,fan2020camouflaged,yang2021uncertainty,lv2021simultaneously}, the combined training set from CAMO and COD10K is used as the training set, and others are test sets.

\noindent\textbf{Metrics.} Following \cite{fan2020camouflaged,le2019anabranch,zhai2021mutual}, four standard metrics are used to comprehensively evaluate the model performance: mean absolute error (MAE), mean E-measure ($E_\phi$) \cite{fan2021cognitive}, S-measure $S_\alpha$ \cite{fan2017structure}, and weighted F-measure $F_\beta^w$ \cite{margolin2014evaluate}.

\noindent\textbf{Implementation Details.} 
We implement our model based on PyTorch in AMD Ryzen Threadripper 3990X 2.9GHz CPU and NVIDIA RTX A6000 GPU. 
For the training stage, the resolution of input images is resized to 704$\times$704, which can avoid the loss of high-resolution information to some extent, and no data augmentation is used in our model. 
The transformer-based feature extraction is initialized by PVT-V2 \cite{Wang_2021_ICCV}, 
and the remaining modules are initialized in a random manner. 
We employ the AdamW~\cite{loshchilov2017decoupled} optimizer with the learning rate of 1e-4, which is widely used in transformer structure, and the corresponding decay rate to 0.1 for every 30 epochs. 
The weight $w$ of iterative feedback loss is 0.2, and the well-optimized iteration number ($in$) is 4. The total epochs of training are 100 with a batch size of 16. For testing, the images are resized to 704$\times$704 as the network's input, and the outputs are resized back to the original size.

\noindent\textbf{Competitors.}
We compare our \ourmodel~with recent 29 state-of-the-art (SOTA) methods, including the most recent COD, salient object detection, generic object detection, and semantic segmentation methods (Tab.~\ref{tab:ModelScore}). 
For a fair comparison, all results are either provided by the published paper or reproduced by an open-source model re-trained on the same training set with the recommended setting. 

\begin{figure*}[t!]
	\centering
    \small
	\begin{overpic}[width=.98\textwidth]{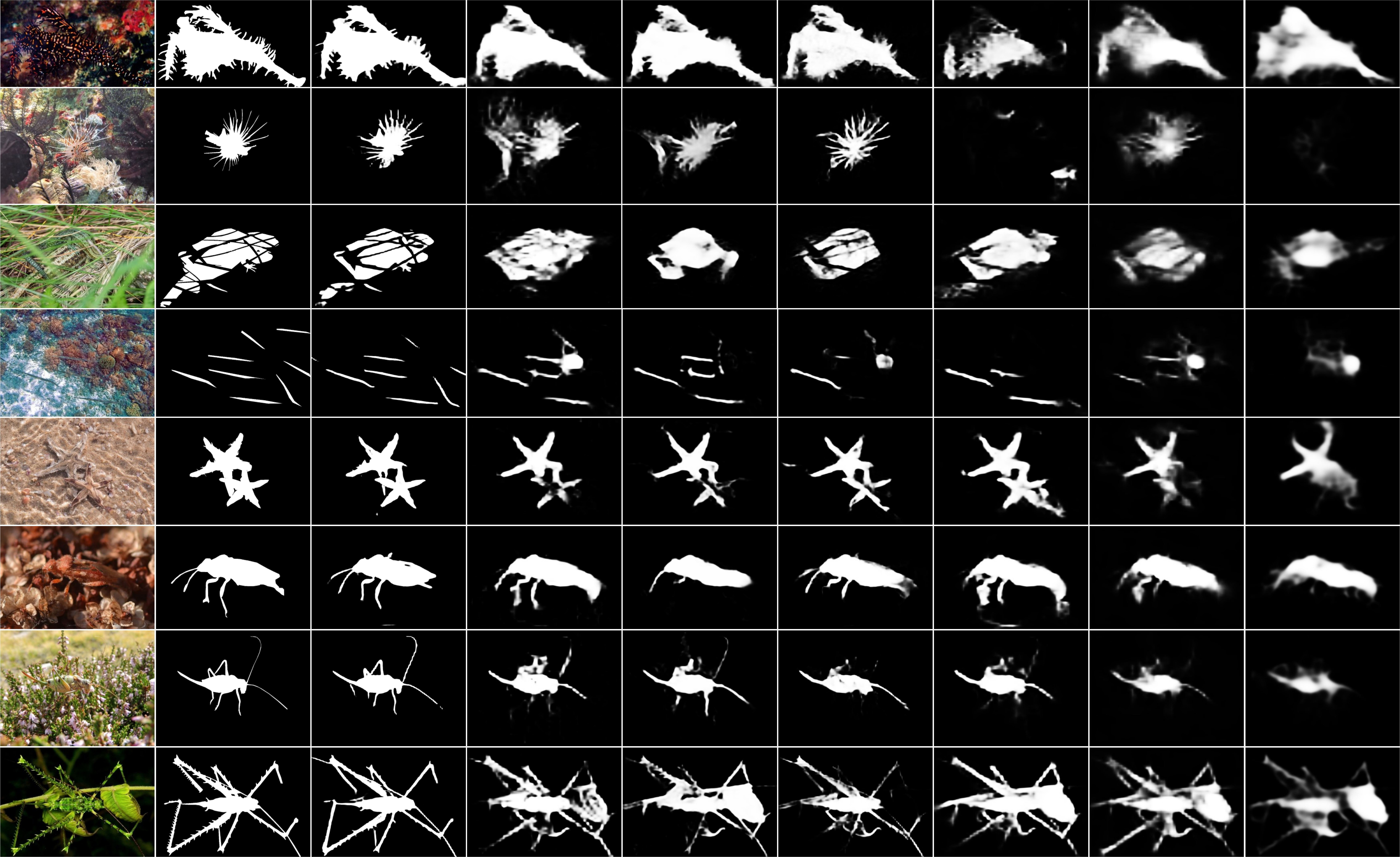} 
    \put( 2,-2){\scriptsize  (a) Images}
    \put(14.5,-2){\scriptsize (b) GT}
    \put(23.5,-2){\scriptsize (c) Our \ourmodel}
    \put(34,-2){\scriptsize (d) SINet-V2 \cite{fangtpami2021}}
    \put(46,-2){\scriptsize (e) PFNet \cite{mei2021camouflaged}}
    \put(57.5,-2){\scriptsize (f) LSR \cite{lv2021simultaneously}}
    \put(68,-2){\scriptsize (g) PraNet \cite{fan2020pranet} }
    \put(79,-2){\scriptsize (h) SINet \cite{fan2020camouflaged}}
    \put(90,-2){\scriptsize (i) EGNet \cite{zhao2019EGNet}}
    \end{overpic}
	\caption{Visual performance of the proposed \textit{\ourmodel}. Our algorithm is capable of tackling  challenging cases (\textit{e.g.,} complex edges with dense thorn, multiply camouflaged objects, partly occlusion, and global thin edges).}
    \label{fig:fig4_visual}
\end{figure*}
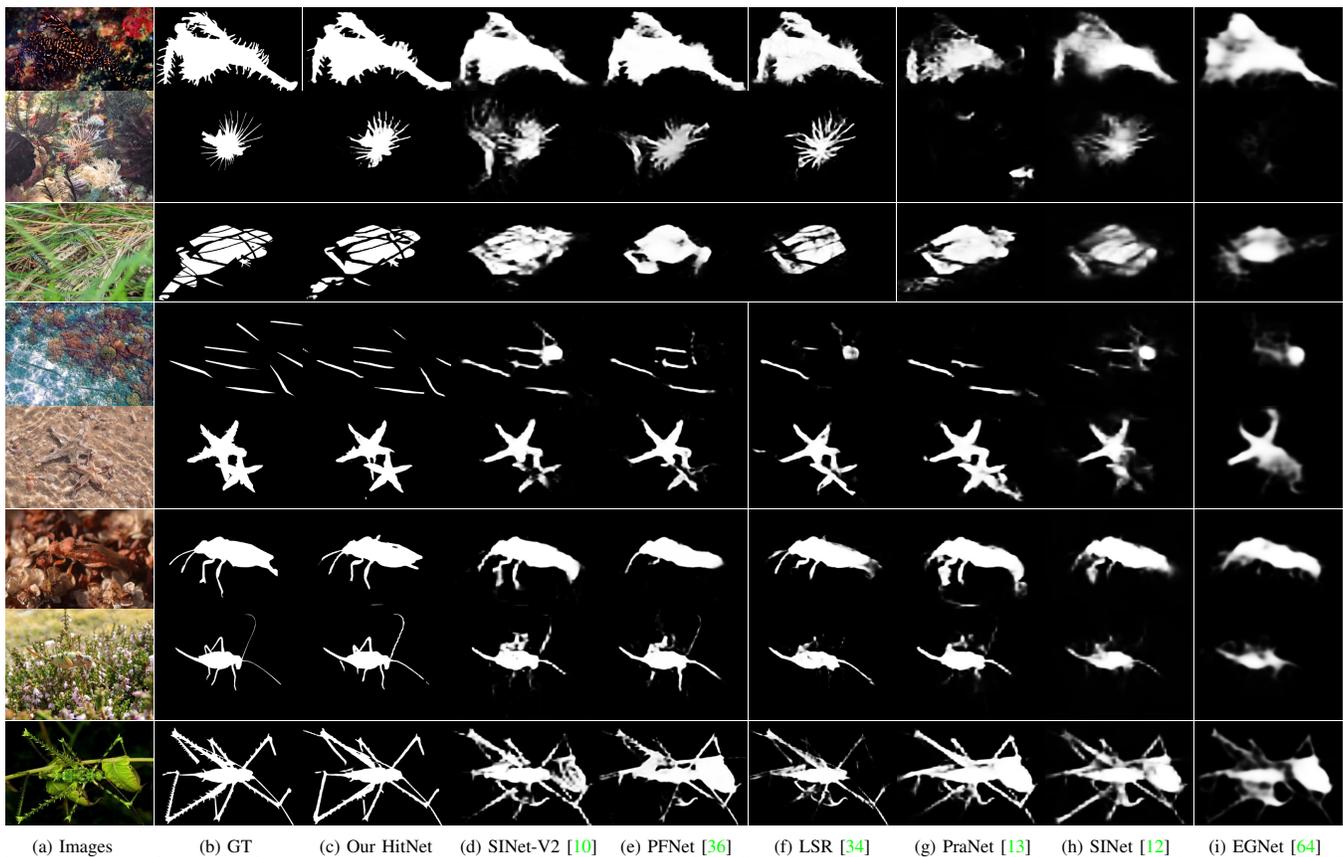

\subsection{Quantitative Evaluation}
\noindent\textbf{CHAMELEON.} As shown in \tabref{tab:ModelScore}, we compare our \textit{\ourmodel} against 27 SOTA algorithms on four standard metrics. As a performance milestone, compared with {second-best} models \cite{fangtpami2021, zhai2021mutual, lv2021simultaneously}, our \textit{\ourmodel} significantly lowers the MAE error by \textbf{40.0\%} and improves $F_\beta^w$ by \textbf{7.6\%}.

\noindent\textbf{CAMO.} For CAMO dataset, compared with 29 models, our \textit{\ourmodel} still dramatically reduces the MAE error by \textbf{18.6\%} and increases $F_\beta^w$ by \textbf{7.8\%} in contrast to second-best~\cite{fangtpami2021}.

\noindent\textbf{NC4K.} We evaluate the generalization ability of all models on NC4K dataset. From Tab. \ref{tab:ModelScore}, as the best model compared with second-best models \cite{fangtpami2021, lv2021simultaneously}, our algorithm reduces the MAE error by \textbf{18.7\%} and improve $F_\beta^w$ by \textbf{5.9\%}.

\noindent\textbf{COD10K.} \tabref{tab:ModelScore} also compares our \textit{\ourmodel} with other 27 SOTA models on the most challenging COD10K test set. From the comparisons, our \textit{\ourmodel} sets a remarkable record to decrease the MAE error by \textbf{31.4\%} and improve $F_\beta^w$ by \textbf{16.5\%} than the second-best models \cite{lv2021simultaneously, yang2021uncertainty, zhai2021mutual}. 
The performance superiority of \textit{\ourmodel} on four benchmarks is mainly due to the well-exploited high-resolution information and the mitigating of high-resolution degradation at the feature level via an iterative feedback mechanism. 

\begin{figure}[t!]
\centering
\includegraphics[width=0.5\textwidth]{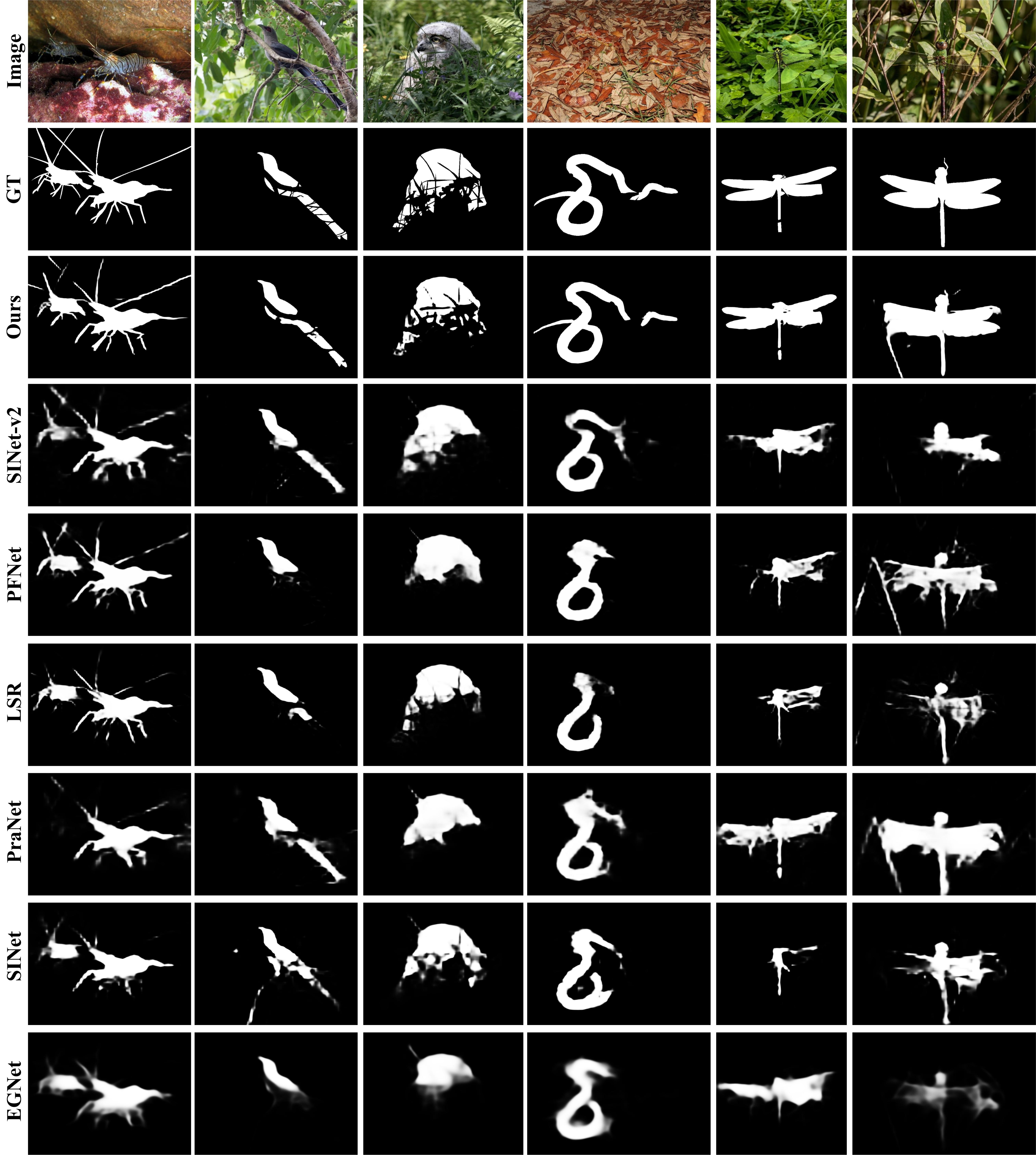}
\caption{Visual performance of the proposed HitNet comparison with state-of-the-art methods (SINet-V2 \cite{fangtpami2021}, PFNet \cite{mei2021camouflaged}, LSR \cite{lv2021simultaneously}, PraNet \cite{fan2020pranet}, SINet \cite{fan2020camouflaged}, EGNet \cite{zhao2019EGNet}) on NC4K dataset.
From the left to right columns, the names of images are \textit{54}, \textit{141}, \textit{161}, \textit{201}, \textit{597}, and \textit{601}, respectively.
}\label{fig:fig_NC4K}
\end{figure}

\begin{figure}[t!]
\centering
\includegraphics[width=0.5\textwidth]{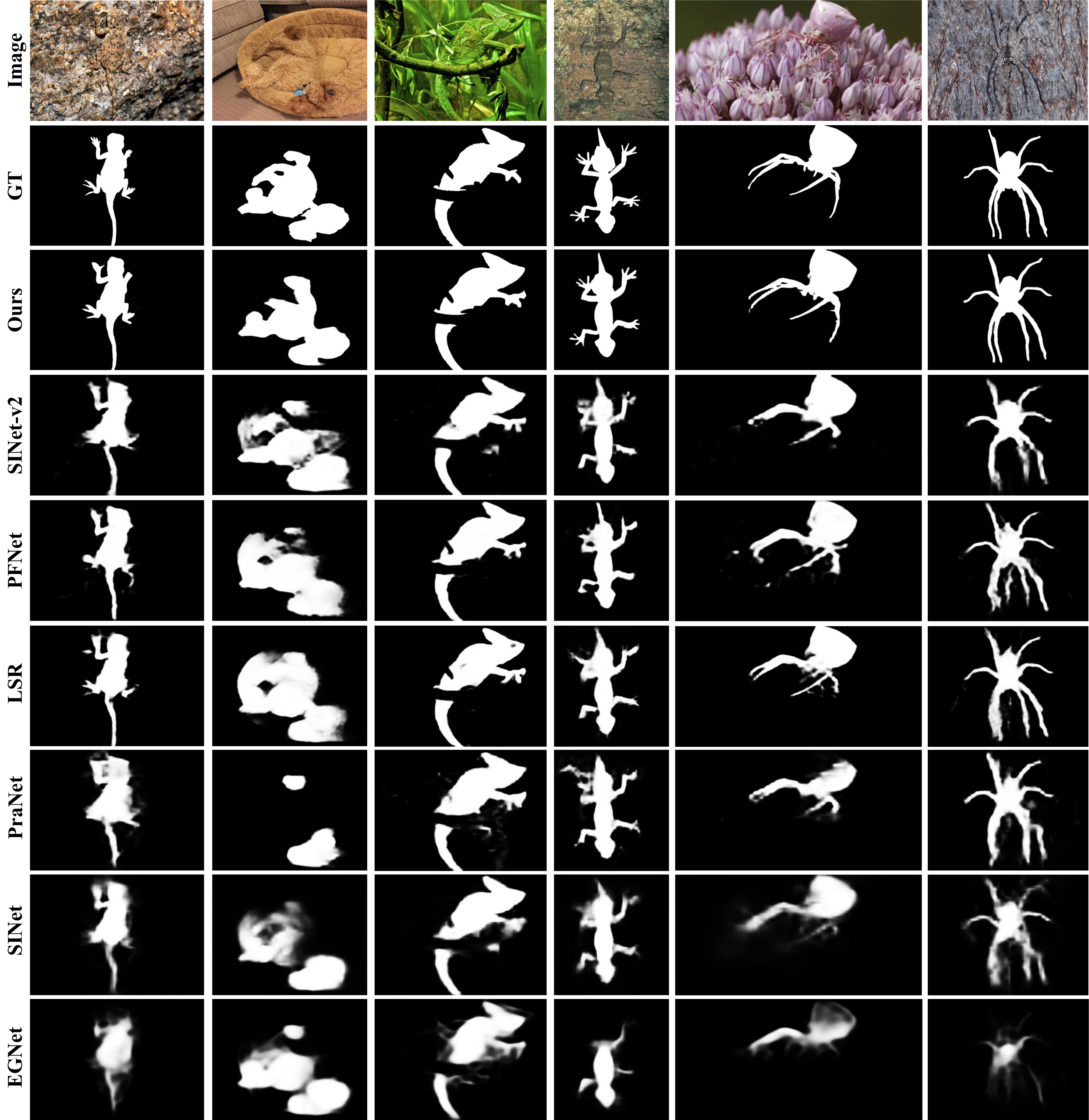}
\caption{Visual performance of the proposed HitNet comparison with state-of-the-art methods (SINet-V2 \cite{fangtpami2021}, PFNet \cite{mei2021camouflaged}, LSR \cite{lv2021simultaneously}, PraNet \cite{fan2020pranet}, SINet \cite{fan2020camouflaged}, EGNet \cite{zhao2019EGNet}) on CHAMELEON dataset. From the left to right columns, the names of images are \textit{animal-9}, \textit{animal-7}, \textit{animal-23}, \textit{animal-33}, \textit{animal-72}, and \textit{animal-70}, respectively.
}\label{fig:fig_CHAMELEON}
\end{figure}

\begin{figure}[t!]
\centering
\includegraphics[width=0.5\textwidth]{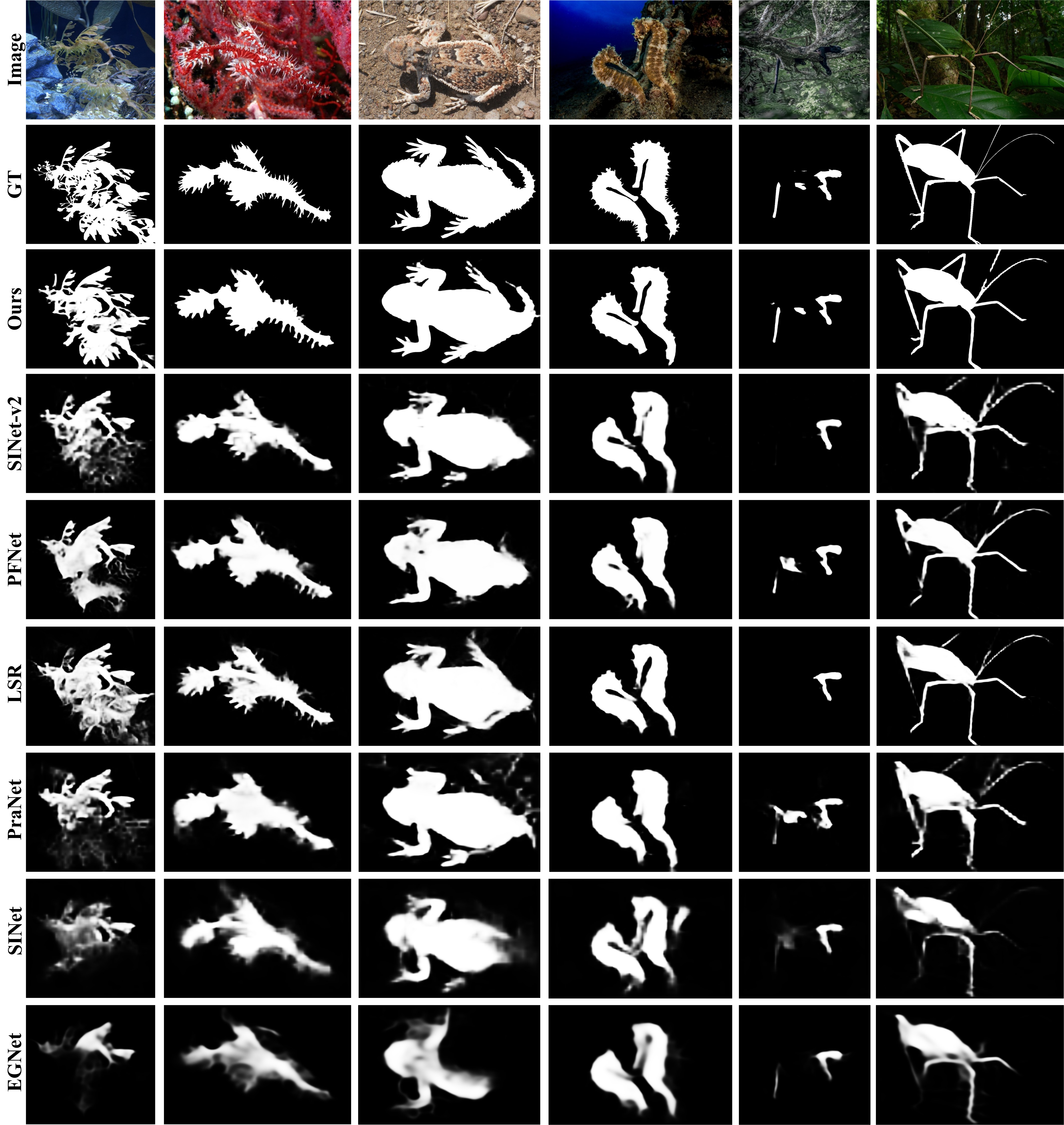}
\caption{Visual performance of the proposed HitNet comparison with state-of-the-art methods (SINet-V2 \cite{fangtpami2021}, PFNet \cite{mei2021camouflaged}, LSR \cite{lv2021simultaneously}, PraNet \cite{fan2020pranet}, SINet \cite{fan2020camouflaged}, EGNet \cite{zhao2019EGNet}) on COD10K dataset.
From the left to right columns, the names of images are \textit{COD10K-CAM-1-Aquatic-10-LeafySeaDragon-423}, \textit{COD10K-CAM-1-Aquatic-9-GhostPipefish-350}, \textit{COD10K-CAM-2-Terrestrial-38-Lizard-2166}, \textit{COD10K-CAM-1-Aquatic-15-SeaHorse-1086}, \textit{COD10K-CAM-5-Other-69-Other-5048}, and \textit{COD10K-CAM-3-Flying-61-Katydid-4196}, respectively.
}\label{fig:fig_COD10K}
\end{figure}

\begin{figure}[t!]
\centering
\includegraphics[width=0.5\textwidth]{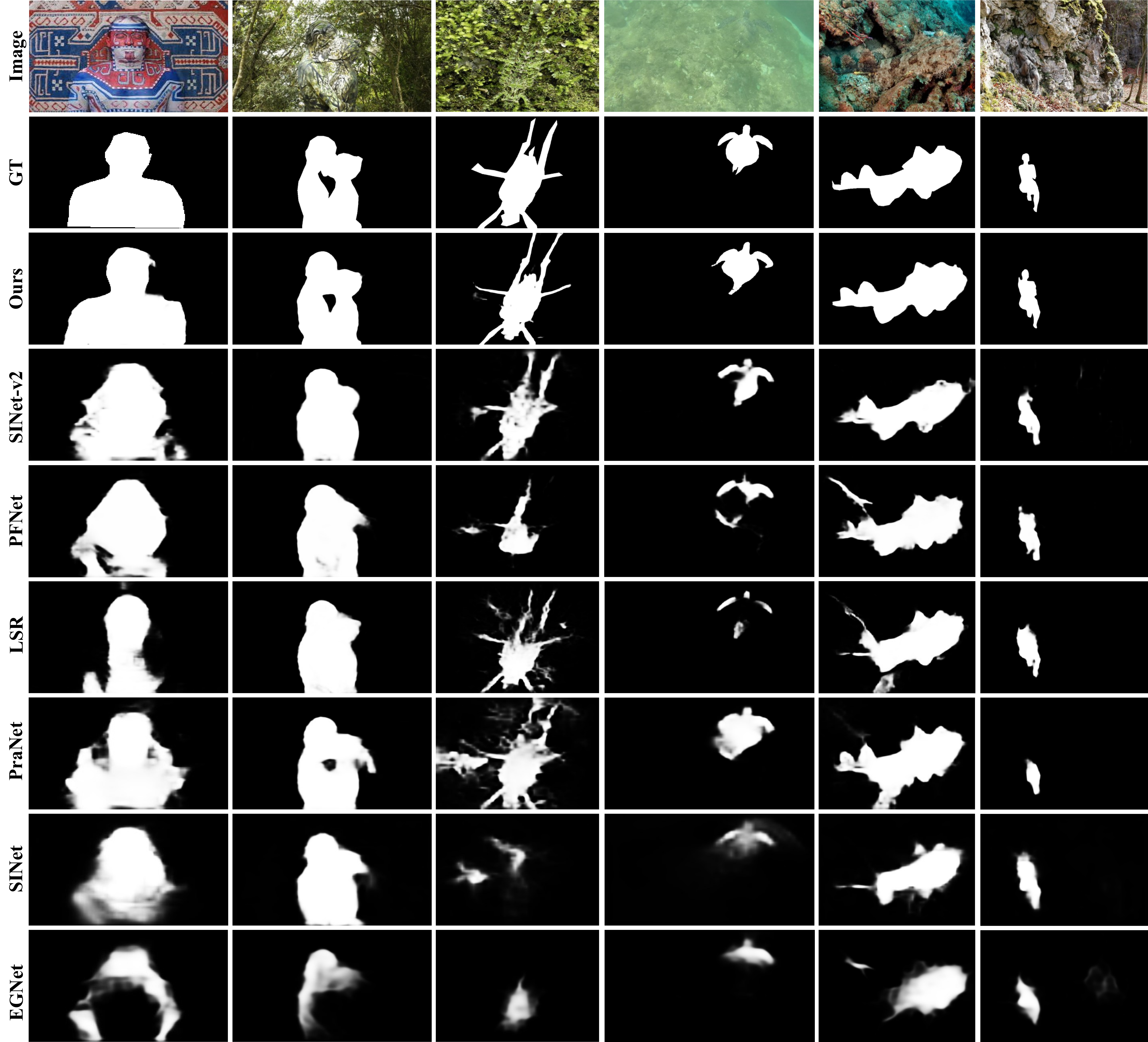}
\caption{Visual performance of the proposed HitNet comparison with state-of-the-art methods (SINet-V2 \cite{fangtpami2021}, PFNet \cite{mei2021camouflaged}, LSR \cite{lv2021simultaneously}, PraNet \cite{fan2020pranet}, SINet \cite{fan2020camouflaged}, EGNet \cite{zhao2019EGNet}) on CAMO dataset. From the left to right columns, the names of images are \textit{camourflage-01181}, \textit{camourflage-01240}, \textit{camourflage-00064}, \textit{camourflage-00135}, \textit{camourflage-00449}, and \textit{camourflage-01243}, respectively.
}\label{fig:fig_camo}
\end{figure}

\subsection{Qualitative Evaluation}
\figref{fig:fig4_visual} shows qualitative results of our \textit{\ourmodel} and other most recent models. 
The examples are difficult to be segmented even for manual annotation due to their complex topological structures and detailed edges from the first row to the third row. But our \textit{\ourmodel} is capable of segmenting clear edges and boundaries (\textit{e.g.,} leaves, thorn) even for objects with occlusion (3-\textit{nd} row). At the same time, other results are blurred or without correct details. For 4-\textit{th} and 5-\textit{th} rows, our \ourmodel~can still clearly segment the multiply camouflaged objects significantly better than others. 
Similar examples from 6-\textit{th} to 8-\textit{th} row also show that 
our algorithm can segment high-resolution edges and do not ignore thin edges (\textit{e.g.,} the antenna or legs of insects) while other models fail in this kind of hard case.
As shown from Fig. \ref{fig:fig_NC4K} to Fig. \ref{fig:fig_camo},
we add more qualitative comparisons with recent state-of-the-art algorithms (2021 SINet-V2 \cite{fangtpami2021}, 2021 PFNet \cite{mei2021camouflaged}, 2021 LSR \cite{lv2021simultaneously}, 2020 PraNet \cite{fan2020pranet}, 2020 SINet \cite{fan2020camouflaged}, 2019 EGNet \cite{zhao2019EGNet}) on four benchmarks (CAMO, CHAMELEON, NC4K, COD10K).

\begin{figure*}[t!]
\centering
\includegraphics[width=0.96\textwidth]{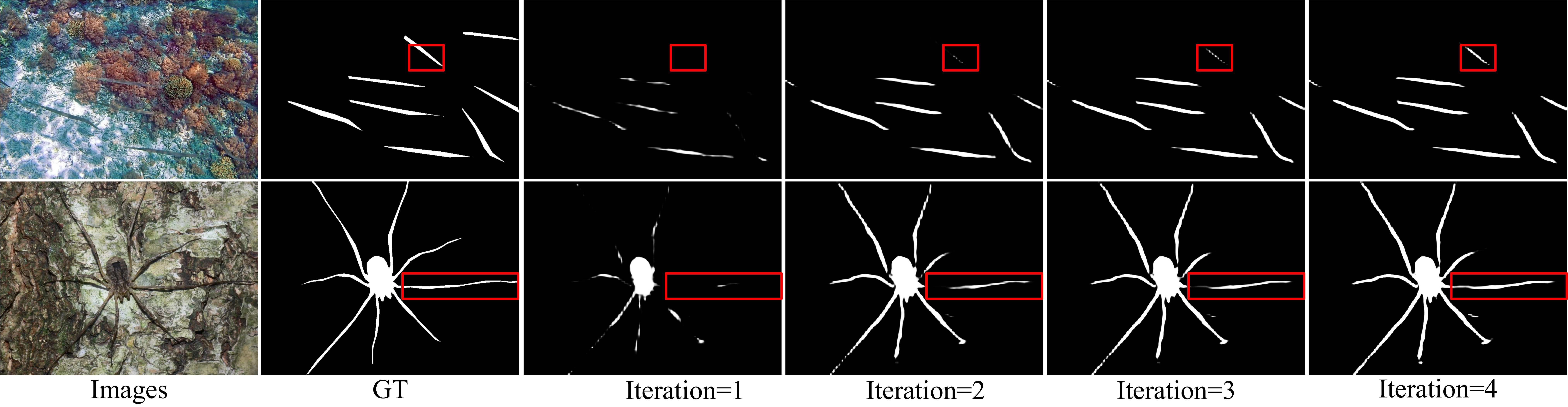}
\caption{Visual performance of each iteration in our iterative feedback mechanism in our RIR module (\secref{section:3parts}).
}\label{fig:fig6_itnum}
\end{figure*}

\subsection{Ablation Study}
\noindent\textbf{Effectiveness of Each Component.}
As shown in \tabref{tab:tab3_3modules}, we evaluate the effectiveness of each module by removing the corresponding part from our complete (\textit{i.e.,} TFE + RIR + IFF) \ourmodel. To assess the contribution of the transformer backbone, we substitute the transformer backbone with Res2Net-50~\cite{gao2019res2net} used in SINet-V2 \cite{fangtpami2021} as the version of `w/o TFE'. It is trained and then evaluated on the most challenging COD10K dataset to show its importance. 
Our algorithm without PVT backbone still achieves the best performance compared with all 29 SOTA methods. But compared with the Res2Net-50 backbone, PVT can achieve better performance due to its superiority of the global receptive field. Besides, we also remove the RIR module from \ourmodel~expressed as `w/o RIR'. The $F_\beta^w$ performance of this variant sharply deteriorates from 0.798 to 0.703. Lastly, we also replace the Iteration Feature Feedback (IFF) with a convolutional fusion layer (`w/o IFF') and find this variant also decreases the performance to some extent. Overall, our RIR module plays a crucial role in performance improvement compared with the other two modules.

\noindent\textbf{Configuration of Iteration Number.}
We explore the effect of iteration number in the iterative feedback mechanism on inference time and performance. As shown in \tabref{tab:tab2_iterativenum}, the performance is gradually improved when the iterative number ($in$) increases from 1 to 5. 
To balance inference time and performance, we choose $T=4$ as default in our \ourmodel. Note that our \textit{\ourmodel} is a real-time algorithm (39 fps), and it sets a new record that is significantly better than 29 other models on four benchmarks.

\noindent\textbf{Study of Iterative Feedback Mechanism.}
As discussed in \secref{section:3parts}, three components enable the feedback mechanism to boost the performance: 
1) Tie each iteration with loss (denoted as `Tie'); 
2) The Feedback Block to avoid the loss of high-resolution information (denoted as `FB');
3) multi-scale connection fusion (denoted as `Multi-fusion'); As shown in Table \ref{tab:tab3_3parts}, any absence of three factors will fail the model to drive the data flow. 
To analyze the difference among the iteration number, we visualize the feature of each iteration in \figref{fig:fig6_itnum}. We observe that the iterative feedback mechanism is a self-correcting process that the subsequent iterations can generate better representations than the previous iteration (\textit{e.g.,} sharper edges).

\noindent\textbf{Analyses of Input Resolution.}
Preventing the loss of HR knowledge (\textit{e.g.,} boundaries or edges) caused by degrading input images is an efficient strategy for COD. But even given same degraded 352$\times$352 inputs, as shown in \tabref{tab:tab3_resolution}, our \ourmodel~is still the best algorithms and reduce the MAE error by 21.6\% than  second-best SINet-V2 \cite{fangtpami2021}.

\noindent\textbf{Evaluation of Different Backbones.}
To assess the contribution of the CNN-based and Transformer-based backbones, we substitute the Transformer backbone of \ourmodel~with Res2Net-50~\cite{gao2019res2net} used in SINet-V2 \cite{fangtpami2021} as the version of `\ourmodel+Res2Net-50'. In addition, we also replace Transformer backbone with ResNet-50 \cite{he2016deep} used in PFNet \cite{mei2021camouflaged} as the version of 
`\ourmodel+ResNet-50'.
All models are trained and then evaluated on the most challenging COD10K dataset to show its importance.
As shown in Tab. \ref{tab:backbone}, compared with 2021~SINet-V2 \cite{fangtpami2021} and 2021~PFNet \cite{mei2021camouflaged} models with the same backbones as ours, our \ourmodel~achieves superior performance with high quantitative results.
Our algorithm (`\ourmodel+Res2Net-50') without Transformer backbone still achieves the best performance compared with all 29 SOTA methods. %
But compared with the Res2Net-50 and ResNet-50 backbone, Transformer can achieve better performance due to its superiority of the global receptive field.

\noindent{\textbf{Inference time.}}
For inference time analysis without considering I/O time, the batch size is set as 1 with the image resolution 704$\times$704. The test stags are also implemented on PyTorch in AMD Ryzen Threadripper 3990X 2.9GHz CPU and NVIDIA RTX A6000 GPU. We compared our HitNet with the most recent algorithms in Tab. \ref{tab:tab2_inference_speed}. From this comparison, our HitNet achieves the best performance with $F_\beta^w$ 0.798 (11.3\% higher than the second-best 2021 LSR \cite{lv2021simultaneously}). Meanwhile, our HitNet is also the second-fastest algorithm with the real-time property.

\noindent{\textbf{Computational complexity (CC) of different backbones:}}
We conduct the GFLOPs analysis on the input resolution 704$\times$704 using four variants. As seen in Tab. \ref{tab:tabGFLOPS}, PVT is much lower than the ViT and also has similar GFLOPs CC with a convolutional-based backbone (ResNet50, Res2Net50).

\begin{table}[ht!]
\vspace{-10pt}
\caption{
GFLOPs analyses of different backbones on COD10K.
}\label{tab:tabGFLOPS}
  \footnotesize
\centering
\begin{tabular}{r|c|c|c|c}
  \toprule
 &  ResNet50 &  Res2Net50 & ViT  &  PVT   \\ 
 \hline
GFLOPs $\downarrow$ & 48  & 52  & 76 & 54  \\  \bottomrule
\end{tabular}
\vspace{-10pt}
\end{table}

\begin{table}[t!]
  \footnotesize
  \centering
  \caption{\small Quantitative results based on different backbones. }\label{tab:backbone}
  \vspace{-10pt}
   \renewcommand{\arraystretch}{1} 
  \setlength\tabcolsep{4pt} 
\begin{tabular}{r|r||cccc}
  \toprule
\multirow{2}{*}{Backbone} & \multirow{2}{*}{Methods} & \multicolumn{4}{c}{COD10K~\cite{fan2020camouflaged}} \\ \cline{3-6}
 &  &$S_\alpha\uparrow$   &$E_\phi\uparrow$   &$F_\beta^w\uparrow$   &$M\downarrow$ \\ \hline
\multirow{2}{*}{ResNet-50 \cite{he2016deep}} & 2021~PFNet \cite{mei2021camouflaged} & 0.800 &0.868 &0.660& 0.040 \\ \cline{2-6}
 & HitNet+ResNet-50 & \textbf{0.805} & \textbf{0.885} & \textbf{0.688} & \textbf{0.036} \\ \hline
\multirow{2}{*}{Res2Net-50 \cite{gao2019res2net}} &  2021~SINet-V2 \cite{fangtpami2021} & 0.815 & 0.887 &0.680 &0.037  \\ \cline{2-6}
 & HitNet+Res2Net-50 & \textbf{0.835} & \textbf{0.898} & \textbf{0.735} & \textbf{0.029} \\ \hline
 \rowcolor{mygray}
 PVT  \cite{Wang_2021_ICCV}  & HitNet+PVT & \textbf{0.868} & \textbf{0.932} & \textbf{0.798} & \textbf{0.024 }     \\ 
 \bottomrule
\end{tabular}
\end{table}


\begin{table}[t!]
\footnotesize
\centering
\caption{Ablation analyses of our \ourmodel~on COD10K dataset \cite{fan2020camouflaged}.}\label{tab:tab3_3modules}
\setlength\tabcolsep{10.5pt}
\begin{tabular}{r|c|c|c|c}
\toprule
 Metric & w/o TFE & w/o RIR & w/o IFF & \ourmodel \\  \hline
$F_\beta^w\uparrow$  & 0.735& 0.703 & 0.791 & \textbf{0.798}  \\ 
$S_\alpha\uparrow$   & 0.835 & 0.821 & 0.863 & \textbf{0.868} \\    
$M\downarrow$   & 0.029 & 0.034 & 0.025 & \textbf{0.024} \\    
$E_\phi\uparrow$   & 0.898 & 0.896 & 0.926 & \textbf{0.932} \\    
\bottomrule
\end{tabular}
\end{table}

\begin{table}[t!]
\caption{\small 
Iterative number ($in$) analyses on COD10K dataset. 
}\label{tab:tab2_iterativenum}
\footnotesize
\centering
\begin{tabular}{r|c|c|c|c|c}
\toprule
& $in$=1 & $in$=2 & $in$=3 & $in$=4 & $in$=5 \\ \hline
MAE $\downarrow$ & 0.0252 & 0.0248 & 0.0249 & \textbf{0.0240} &  0.0241\\ 
Test time (\textit{ms}) $\downarrow$ & \textbf{18.8} & 21.1 & 23.4 & 25.6 &  27.6\\ 
\bottomrule
\end{tabular}
\end{table}

\begin{table}[ht!]
\vspace{-10pt}
\caption{Inference speed (\textit{fps}) analyses on COD10K dataset. 
}\label{tab:tab2_inference_speed}
\footnotesize
\centering
\begin{tabular}{r|c|c|c|c|c}
\toprule
 &  \cite{fan2020camouflaged} &  \cite{mei2021camouflaged} &   \cite{lv2021simultaneously} &  \cite{fangtpami2021}  & Ours \\ \hline
$F_\beta^w$  $\uparrow$  & 0.551 & 0.660 & 0.685  & 0.680&  \textbf{0.798}\\ 
Test time (\textit{fps}) $\downarrow$ & 36 & \textbf{40}  & 29 & 31 & {39} \\ 
\bottomrule
\end{tabular}
\vspace{-10pt}
\end{table}

\begin{table}[t!]
\footnotesize
\caption{Ablation study on indispensable factors of Iterative Feedback  Mechanism on COD10K dataset.} \label{tab:tab3_3parts}
\centering
\setlength\tabcolsep{16.5pt}
\begin{tabular}{c|c|c|c}
\toprule
\multicolumn{3}{c|}{Configurations}  & Performance  \\ \hline 
Tie  &  FB &  Multi-fusion & MAE $\downarrow$ \\ 
\hline
$\times$    & $\times$ & $\times$ & 0.0268 \\ 
$\surd$     & $\times$ & $\times$ & 0.0255 \\ 
$\surd$     & $\surd$  & $\times$ & 0.0248 \\ 
$\surd$     & $\surd$  & $\surd$ & \textbf{0.0240} \\ 
\bottomrule
\end{tabular}
\end{table}

\begin{table}[t!]
\footnotesize
\centering
\caption{Input resolution analyses of \ourmodel~on COD10K dataset.}\label{tab:tab3_resolution}
\setlength\tabcolsep{11pt}
\begin{tabular}{c|c|c|c|c}
\toprule
 Resolution &$S_\alpha\uparrow$      &$E_\phi\uparrow$     &$F_\beta^w\uparrow$      &$M\downarrow$\\  \hline
Input: 352$\times$352  & 0.827&  0.907&  0.727&   0.029 \\  
Input: 704$\times$704  & \textbf{0.868} & \textbf{0.932} & \textbf{0.798}  & \textbf{0.024} \\
\bottomrule
\end{tabular}
\end{table}

\section{Application}
As studied by Fan \textit{et al.} \cite{fan2020camouflaged}, the term ``salient'' is essentially the opposite of ``camouflaged''. We are interested in implementing an application that converts salient objects to camouflage objects. We adopt a cross-domain learning (CDL) technique to achieve this goal. In addition, we propose a contrastive index to evaluate the camouflaged level. This index can be acted as the criterion to discard some hard cases with unchangeable intrinsic salient objects.

\begin{figure}[t!]
\centering
\includegraphics[width=\columnwidth]{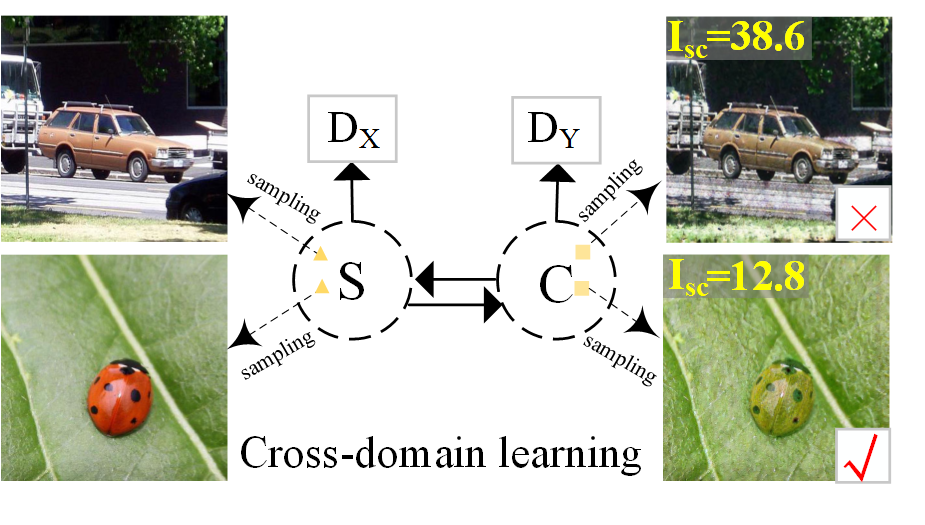}
\caption{The overview of salient-to-camouflaged cross-domain learning pipeline. The $S$ is the salient domain, and the $C$ means the camouflaged domain. $D_{X}$ is the discriminator of the salient domain, and $D_{Y}$ is the discriminator of the camouflaged domain.
}\label{fig:cross_doamin_s}
\end{figure}

\subsection{Cross-domain Learning}\label{sec:CDL}
We employ the cycle-consistency structure~\cite{zhu2017unpaired} to learn the camouflaged features and embed these features into the salient objects in an unsupervised cross-domain learning manner as shown in  \figref{fig:cross_doamin_s}. 
The cycle-consistency loss can be formulated as:
\begin{equation}
\begin{split}
\mathcal{L}_{cyc}(G,F)  = & \mathbb{E}_x[\Arrowvert(F(G(x))-x) \Arrowvert _1] +   \\
 & \mathbb{E}_y  [\Arrowvert (G(F(y))-y) \Arrowvert _1]
\end{split}
\end{equation}
where $G$ aims to construct fake images $\{G(x)\}$ from salient samples $\{x\}$ to get close to camouflaged domain $Y$ while $D(Y)$ tries to distinguish between the translated camouflaged samples $\{G(x)\}$ and real camouflaged samples $\{y\}$. $F$ is another translator from camouflaged to salient objects.  The procedure is concluded as a min-max optimization task\footnote{Minimize the generator loss while maximized the discriminator loss.} in the adversarial loss function used in CycleGAN~\cite{zhu2017unpaired}.

\begin{figure}[t!]
\centering
\includegraphics[width=0.47\textwidth]{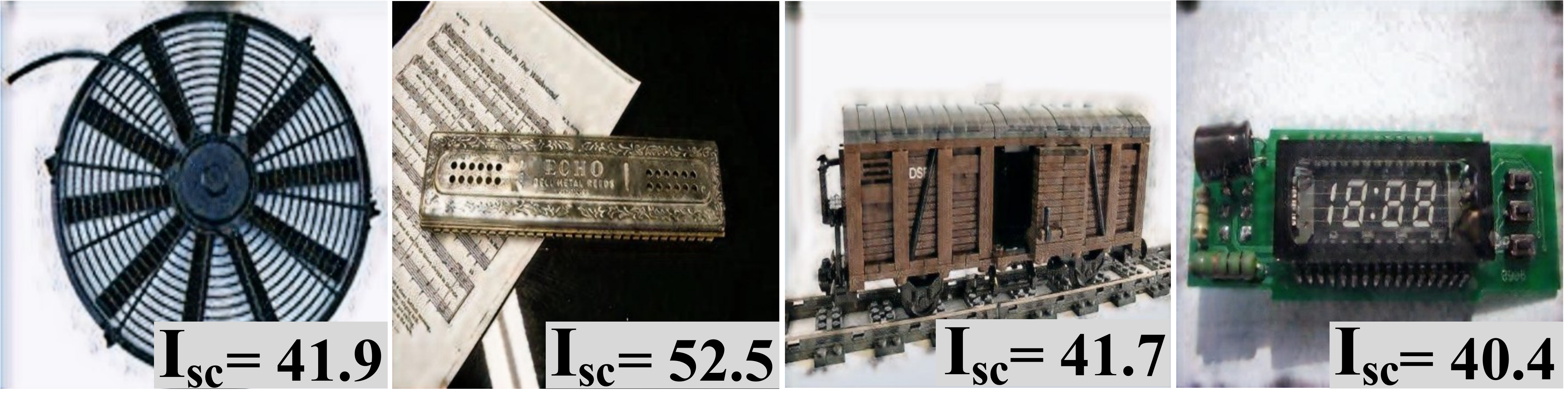}
\caption{Failure cases of CDL detected by our contrastive index.
}\label{fig:fig_failure_case}
\end{figure}

To better select the converted camouflaged objects, we propose a contrastive index, considering the pixel-level similarity between object and its surroundings:
\begin{equation}
{I}_{sc}=\frac{1}{\rm Num} \sum_{i}^{\rm Num} \left \| P_{i} - P_{\rm m} \right\|_{i \in (P_{\rm m}-P_{\rm std},P_{\rm m}+P_{\rm std})},
\end{equation}
where ${I}_{sc}$ is the index of camouflaged level, $P_i$ is $i$-th pixel intensity value, $P_m$ is the mean value of images, $P_{\rm std}$ is the standard deviation, and $i$ is the pixel index that belongs to one $\sigma$ rule to exclude the effect of extreme values. As shown in \figref{fig:cross_doamin_s}, the car is an abandoned example that can be detected as a high salient case by our contrastive index. Empirically, we set the threshold of $I_{sc}$ between the camouflaged and salient object as $I_{sc}=20$ after plenty of observations. With this index, we can discard some failure converted cases, as shown in \figref{fig:fig_failure_case}.

\begin{figure}[t!]
\centering
\includegraphics[width=0.47\textwidth]{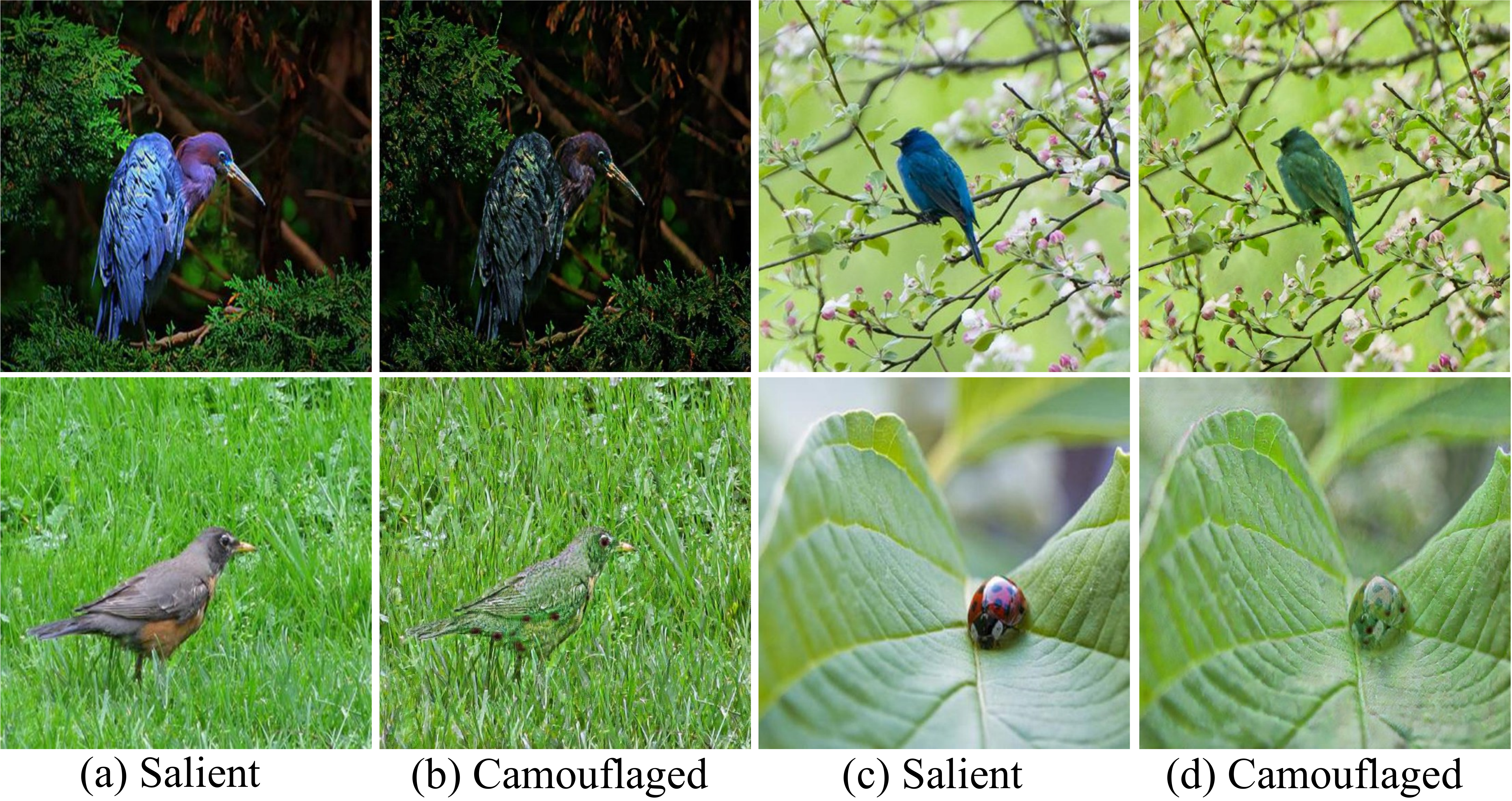}
\caption{Application Results of convert salient object (\textit{i.e.,} a \& c) to camouflaged object (\textit{i.e.,} c \& d).
}\label{fig:fig5_s2c}
\end{figure}

\noindent\textbf{Qualitative and Quantitative Evaluation.}
As shown in \figref{fig:fig5_s2c}, our Cross-Domain Learning (CDL) method can merge the salient objects into the background as camouflaged objects.
As shown in \tabref{tab:tables2c}, although there exist numerous salient object data, it fails to boost the performance in camouflaged scenarios directly. Instead, they severely deteriorate $S_\alpha$ from 0.868 to 0.844 and $E_\phi$ from 0.932 to 0.911. 
Meanwhile, the proposed CDL can improve the $F_\beta^w$ from 0.798 to 0.806 and reduce the MAE error by 4.2\%.

\begin{figure}[t!]
	\centering
    \small
	\begin{overpic}[width=.47\textwidth]{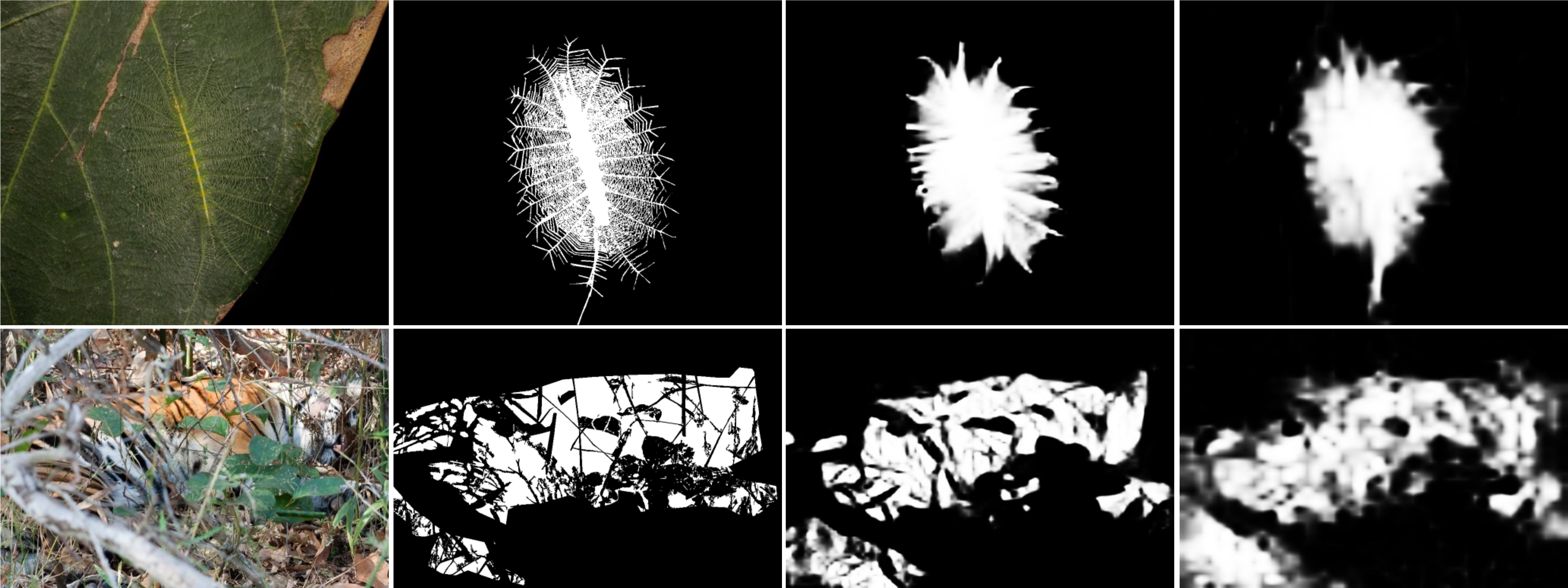} 

    \put( 6,-3){\scriptsize  (a) Images}
    \put(33,-3){\scriptsize (b) GT}
    \put(53,-3){\scriptsize (c) Our \ourmodel}
    \put(78,-3){\scriptsize (d) SINet-V2 \cite{fangtpami2021}}

    \end{overpic}
	\caption{Failure cases of our~\ourmodel.}
    \label{fig:failure_case}
\end{figure}
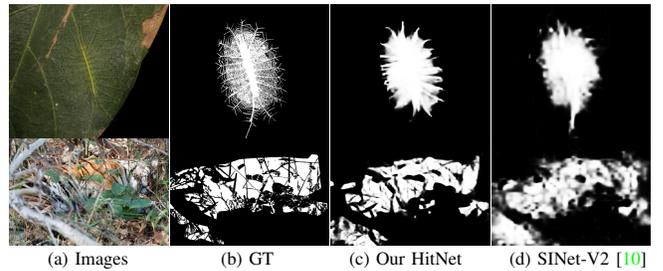

\begin{table}[t!]
  \footnotesize
  \centering
  \caption{\small Quantitative results on different training strategies. `w/o' means without any data strategy, `Salient data' means adding salient data for training.}\label{tab:tables2c}
  \setlength\tabcolsep{11.0pt}
  \begin{tabular}{l|cccc}
  \toprule
  &\multicolumn{4}{c}{\tabincell{c}{COD10K~\cite{fan2020camouflaged}}}  \\
   \cline{2-5}
   Data Strategy~~~ &$S_\alpha\uparrow$      &$E_\phi\uparrow$     &$F_\beta^w\uparrow$      &$M\downarrow$\\ \hline
   w/o &0.868 & 0.932 & 0.798& 0.024 \\ 
   Salient data & 0.844 & 0.911 & 0.782 & 0.026  \\
   \textbf{CDL (Ours)} & \textbf{0.871} & \textbf{0.935} & \textbf{0.806} & \textbf{0.023}   \\
\bottomrule
  \end{tabular}
\end{table}  

\section{Failure Case}
Although our \ourmodel~sets a new record in the COD task, there are still some very challenging examples (\textit{e.g.,} caterpillar) that \ourmodel~fails to address. As shown in \figref{fig:failure_case}, these cases usually contain complicated topological structures with lots of dense edges or details.

\noindent\textbf{Limitation of Multi-resolution Iterative Refinement.} For some very challenging cases, \textit{e.g.,} one single camouflaged object with complex topological structures, global long-range edges, or 
the multiply camouflaged objects where one of objects is very small to be neglected, our algorithm still has some space to be improved. 
The bottleneck of the Iterative Feedback Mechanism mainly causes the potential reasons. This mechanism segments the dense edges or multiplies small objects by self-correcting low-resolution features with high-resolution information. Unfortunately, when the iteration number$>$4, the performance only improves slightly but increases time. 

\section{Conclusion}
We propose a novel high-resolution iterative feedback network (\textbf{\ourmodel}) to extract the informative and high-resolution representations for tackling the degradation issue of segmentation details on the COD task. \ourmodel~can adaptively refine the low-resolution features with high-resolution information in an iterative feedback manner. 
More importantly, our approach achieves remarkable performance improvements and significantly outperforms 29 cutting-edge models on four challenging datasets. Finally, we introduce the cross-domain learning strategy to implement an application that converts the salient object to the camouflaged object, potentially enlarging the diversity of the COD dataset. 

\noindent\textbf{Broader Impacts.} 
As a performance milestone, our HitNet has a great potential to be deployed in real scenarios of camouflaged object detection (\textit{e.g.,} species discovery, helicopter rescue, polyp segmentation). In academia, this work may give some inspiration to explore high-resolution features to facilitate the finer segmentation. Our cross-domain strategy builds a bridge between camouflaged and salient domains, where the camouflaged domain can benefit from the diverse salient dataset to some extent.

{\small
\bibliographystyle{ieee_fullname}
\bibliography{camouflage}
}


 


\vspace{11pt}


\vfill

\end{document}